\documentclass{article}

\usepackage{microtype}
\usepackage{graphicx}
\usepackage{subfigure}
\usepackage{booktabs} 

\usepackage{hyperref}

\usepackage[accepted]{icml2024}

\usepackage{amsmath}
\usepackage{amssymb}
\usepackage{mathtools}
\usepackage{amsthm}

\usepackage[capitalize,noabbrev]{cleveref}

\theoremstyle{plain}

\theoremstyle{definition}

\theoremstyle{remark}

\usepackage[textsize=tiny]{todonotes}

\usepackage{capt-of}
\usepackage{xspace}
\newcommand{\metabbr}{InstructBooth\xspace}

\icmltitlerunning{InstructBooth: Instruction-following Personalized Text-to-Image Generation}

\begin{document}

\twocolumn[{
\icmltitle{InstructBooth: Instruction-following Personalized Text-to-Image Generation}



\icmlsetsymbol{equal}{*}

\begin{icmlauthorlist}
\icmlauthor{Daewon Chae}{korea}
\icmlauthor{Nokyung Park}{korea}
\icmlauthor{Jinkyu Kim}{equal,korea}
\icmlauthor{Kimin Lee}{equal,kaist}
\end{icmlauthorlist}

\icmlaffiliation{korea}{Korea University}
\icmlaffiliation{kaist}{KAIST}

\icmlcorrespondingauthor{Jinkyu Kim}{jinkyukim@korea.ac.kr}
\icmlcorrespondingauthor{Kimin Lee}{kiminlee@kaist.ac.kr}

\icmlkeywords{Machine Learning, ICML}

\vskip 0.3in

}
]



\printAffiliationsAndNotice{}  

\begin{abstract}
    Personalizing text-to-image models using a limited set of images for a specific object has been explored in subject-specific image generation. However, existing methods often face challenges in aligning with text prompts due to overfitting to the limited training images. In this work, we introduce \metabbr, a novel method designed to enhance image-text alignment in personalized text-to-image models without sacrificing the personalization ability. Our approach first personalizes text-to-image models with a small number of subject-specific images using a unique identifier. After personalization, we fine-tune personalized text-to-image models using reinforcement learning to maximize a reward that quantifies image-text alignment. 
Additionally, we propose complementary techniques to increase the synergy between these two processes. Our method demonstrates superior image-text alignment compared to existing baselines, while maintaining high personalization ability. In human evaluations, \metabbr outperforms them when considering all comprehensive factors. Our project page is at \href{https://sites.google.com/view/instructbooth}{https://sites.google.com/view/instructbooth}.

\begin{figure}[t]
    \centering
    \centerline{\includegraphics[width=\columnwidth]{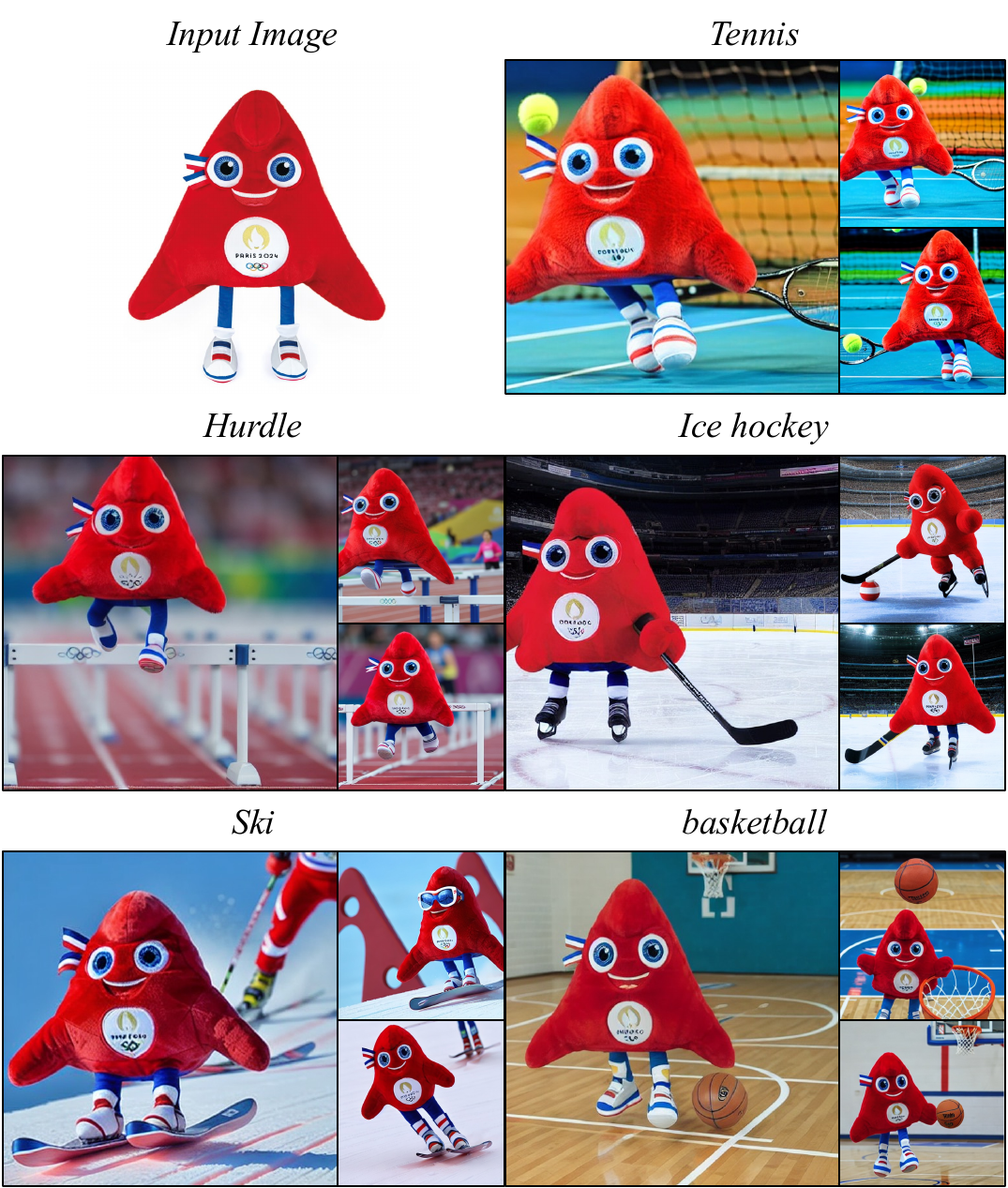}}
    \vspace{-1em}
    \caption{We propose \metabbr, a method that enables the generation of images featuring specific user-provided subjects (such as characters or objects) based on a few input reference images without degradation in image-text alignment. For example, \metabbr can create new images of {\em unseen} Phryge, the Paris 2024 Olympic mascot plushie, participating in various sports.}
    \label{fig:phryge_main}
    \vspace{-1em}
\end{figure}
\end{abstract}

\section{Introduction}

Recently, text-to-image models \cite{dalle2,imagen,sd} have demonstrated superior performance in generating natural and high-quality images given novel text prompts. These models can produce photorealistic images of general objects in diverse contexts using a natural language prompt. Based on these advancements, the new research question arises: 
{\em How can we enable text-to-image models to generate personalized subject images?}
For example, given only a few images of an Olympic mascot plushie, the goal is to make models generate images featuring this plushie in various contexts, such as participating in Olympic sports at the stadium (as illustrated in Figure~\ref{fig:phryge_main}). This capability holds the potential to open up exciting possibilities in personalized image generation, empowering users to effortlessly create custom imagery tailored to their specific interests and preferences.

To personalize existing text-to-image models, several approaches have been proposed that learn user-defined concepts using a few given images \cite{dreambooth, textual_inversion, P+, neti, prospect,cones}. 
For example, \citet{dreambooth} proposed DreamBooth, a method that fine-tunes pre-trained models using a unique identifier. 
Specifically, it trains the text-to-image model using a text prompt formatted as “[identifier] [class noun]” and applies regularization techniques using the dataset of the class to which the subject belongs.
This approach demonstrates a high personalization ability.

\begin{figure}[t]
    \centering
    \includegraphics[width=1.0\linewidth]{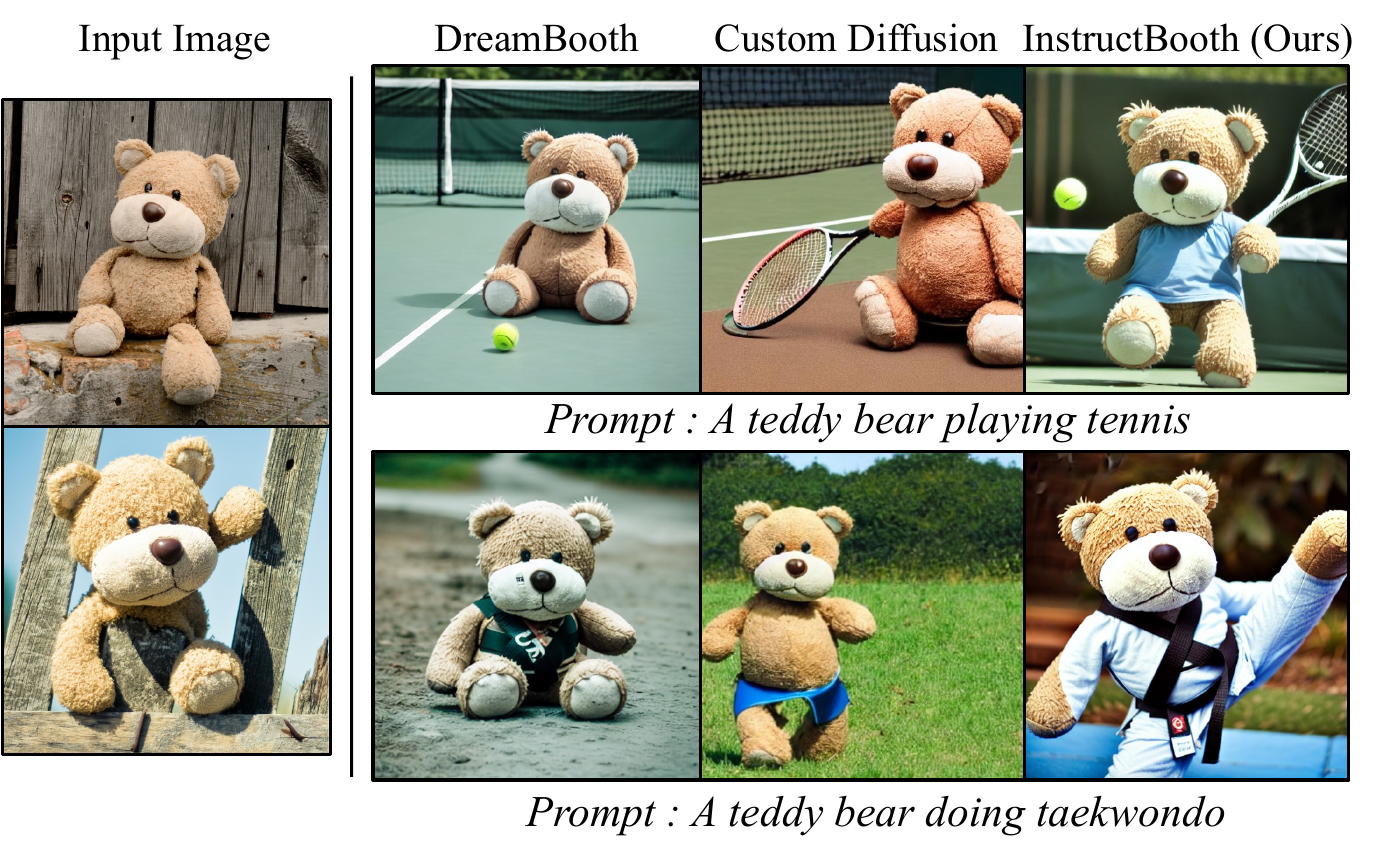}
    \vspace{-2em}
    \caption{Comparison of images generated by DreamBooth, Custom Diffusion and \metabbr with a few images of a specific object (left) on given text prompt.}
   \label{fig:comparison}
   \vspace{-1em}
\end{figure}

Despite the promise of prior methods, they often exhibit issues with low text fidelity due to overfitting on the limited training images.
As shown in Figure~\ref{fig:comparison}, when provided with the text prompt ``playing tennis'' or ``doing taekwondo'', DreamBooth \cite{dreambooth} becomes overfitted and fails to reflect the desired action which input prompt requires. To mitigate this overfitting issue, several studies \cite{multiconcept, svdiff, key_locked} have proposed the personalization method that constrains the trainable weights of pre-trained text-to-image models. While this approach shows some improvement in text fidelity, it often results in significant degradation of personalization capability. In the case of Custom Diffusion~\cite{multiconcept}, which only fine-tunes specific layers, the generated images are notably less similar to the reference images compared to those produced by DreamBooth, as demonstrated in Figure~\ref{fig:comparison}.
Therefore, given the limitations of recent approaches, enhancing both subject and text fidelity remains a significant challenge in the personalization task.

In this work, we aim to achieve both high subject personalization and text fidelity by addressing the challenges associated with supervised learning using a limited dataset of images. Our main idea is to introduce reinforcement learning (RL) fine-tuning in a subsequent stage. 
Specifically, we first personalize text-to-image model by updating model parameters with a unique identifier and a few reference images similar to DreamBooth~\cite{dreambooth}. After personalization, we employ RL fine-tuning to address potential overfitting issues. During RL fine-tuning, the personalized model generates new images of the subject using prompts designed to ensure the model reflects the desired characteristics faithfully. The model then receives rewards for its outputs based on the image-text alignment. We update the model to maximize the rewards using a policy gradient method~\cite{ddpo,dpok}. This iterative process effectively mitigates overfitting and enables the model to generate subject images with a high level of alignment to the provided text descriptions (see Figure~\ref{fig:comparison}). 

Furthermore, we introduce complementary techniques to enhance the synergy between the two processes. In the first personalization step, in addition to a unique identifier, we include the detailed description in the text prompt for rare subjects (e.g., using ``a [identifier] \textit{triangular} plushie'' instead of ``a [identifier] plushie'' when learning a triangular-shaped mascot like Phryge in Figure~\ref{fig:phryge_main}). 
This approach allows models to capture the subject's characteristics precisely, facilitating personalization even for rare subjects.

In the RL fine-tuning step, we design training text prompts to efficiently improve text fidelity. 
Specifically, we utilize both prompts with and without a unique identifier (i.e., ``[identifier] [class noun] [activity descriptor]'' and ``[class noun] [activity descriptor]'').
For example, while fine-tuning the model to generate images of a Paris Olympic mascot plushie playing tennis using the prompt with an identifier, we simultaneously train the model to produce images of a general plushie playing tennis using the prompt without an identifier.
This method alleviates the difficulty caused by the overfitted personalized model (i.e., starting point of RL fine-tuning), which often struggles to produce a variety of high-quality samples and rarely delivers good reward signals. By combining all proposed techniques, our method can generate personalized images of the Paris 2024 Olympic mascot plushie that align with text prompts (see Figure~\ref{fig:phryge_main}).

We evaluate the performance of our proposed method on a diverse set of subject images, comparing it to prior methods, such as DreamBooth \cite{dreambooth}, Custom Diffusion \cite{multiconcept}, NeTI \cite{neti}, and Textual Inversion \cite{textual_inversion}. In both quantitative and qualitative evaluations, our method outperforms the other methods in terms of image-text alignment, while maintaining high subject fidelity. Additionally, our human evaluation shows that human raters prefer \metabbr over other models in side-by-side comparisons. 
We summarize our contributions as follows:
\vspace{-1em}
\begin{itemize}
    \item 
    We propose \metabbr: a novel personalization approach that effectively mitigates overfitting issues via RL fine-tuning. Our method produces personalized images with high text fidelity while preserving subject fidelity. \vspace{-.5em}
    \item 
    We also investigate important design choices, such as selecting training prompts to improve the efficiency of RL fine-tuning and enhancing the synergy with personalization. \vspace{-.5em} 
    \item 
    We conduct human evaluations to demonstrate the effectiveness of our method compared to current state-of-the-art approaches.
    InstructBooth outperforms all baselines in terms of text fidelity while showing superior perceptual similarity of subjects.
\end{itemize}

\section{Related Work}
\paragraph{Personalized Text-to-Image Generation.}
Since text-to-image models \cite{dalle2,imagen,sd,muse,parti} have shown impressive results, several studies have been proposed to personalize the text-to-image models using only a few images of a specific subject. Specifically, two distinct streams of research have emerged to achieve this goal. The first stream, based on Textual Inversion \cite{textual_inversion}, focuses on optimizing new word embedding to represent a given subject or concept. This line of research demonstrates high controllability and has recently been expanded to learn new word embedding in various embedding spaces \cite{neti, prospect, P+, agarwal2023image}. The second stream, based on DreamBooth \cite{dreambooth}, involves the method to fine-tune the pre-trained models using the text prompt with a unique identifier. Within this line of research, each approach is differentiated by determining the scope of weights to be learned, such as fine-tuning only the weights of the attention layer \cite{multiconcept, key_locked, svdiff, orthogonal}. Our method is based on the latter (i.e., DreamBooth) in terms of personalizing the pre-trained models using a unique identifier. However, unlike prior works, we introduce reinforcement learning (RL) in a subsequent step. This incorporation of RL effectively mitigates overfitting, enabling the generation of images with high text fidelity for prompts that have been challenging for existing methods.

\paragraph{Improving image-text alignment.} Despite the impressive success of text-to-image models, they often struggle to generate images that accurately align with text prompts. 
To address this issue, recent studies have investigated {\em learning from human feedback} in text-to-image generation~\cite{lee2023aligning, imagereward, kirstain2023pick, hps}.
These methods first learn a reward function intended to reflect what humans care about in the task, using human feedback on model outputs. 
Subsequently, they utilize the learned rewards to enhance image-text alignment through techniques such as rejection sampling \cite{kirstain2023pick}, reward-weighted learning \cite{lee2023aligning, hps}, and direct reward optimization via gradient \cite{imagereward, draft, alignprop}. 
DDPO~\cite{ddpo} and DPOK~\cite{dpok} formulate the fine-tuning problem as a multi-step decision-making problem and propose a policy gradient method to maximize the expected rewards.
They demonstrate that RL fine-tuning can effectively improve text-to-image alignment. Inspired by their successes, we propose to utilize the fine-tuning with the reward model to improve the image-text alignment of personalized text-to-image models.

\section{Preliminary}
In this work, we consider text-to-image diffusion models, a class of generative models that transform Gaussian noise via iterative denoising process to model data distribution $p(\mathbf{x})$~\cite{ddpm}.
Specifically, we utilize a latent diffusion model~\cite{sd}, which (i) operates in a highly compressed lower-dimensional
latent space $\mathbf{z}=\mathcal{E}(\mathbf{x})$ using an image encoder $\mathcal{E}$, rather than relying on pixel-based image representations $\mathbf{x}$, and (ii) utilizes a conditional denoising autoencoder $\epsilon_\theta(\mathbf{z}, t, \mathbf{c})$ to control the synthesis process with inputs $\mathbf{c}$ (i.e., language prompts) by modeling conditional distributions $p(\mathbf{z}|\mathbf{c})$. The training of this model involves predicting the added noise $\epsilon$ to the latent representation $\mathbf{z}_t$ at timestep $t$, using an additional condition $\mathbf{c}$. Formally, the objective of the training is as follows:

\begin{equation}
    \mathcal{L}_{LDM} :=\mathbb{E}_{\mathbf{z}, \mathbf{c}, \epsilon, t}\left[\left\|\epsilon-\epsilon_\theta\left(\mathbf{z}_t, t, \mathbf{c}\right)\right\|_2^2\right],
    \label{eq:sd}
\end{equation} where $t$ is timestep uniformly sampled from $\{1, 2, \dots, T\}$, $\epsilon$ is a Gaussian noise $\sim \mathcal{N}(0,I)$, and $\mathbf{z}_t$ is the noised latent representation at a timestep $t$. At inference phase, the iterative denoising process is conducted to produce the denoised sample $\mathbf{z}_0$, using the noise predicted by $\epsilon_\theta(\mathbf{z}_t, t, \mathbf{c})$. Specifically, the predicted noise is used to calculate the mean of transition distribution $p_\theta\left(\mathbf{z}_{t-1} \mid \mathbf{z}_t,\mathbf{c}\right)$ in denoising process. 
The formulation is as follows:
\begin{equation}
\mu_\theta\left(\mathbf{z}_t, t, \mathbf{c}\right)=\frac{1}{\sqrt{\alpha_t}}\left(\mathbf{z}_t-\frac{\beta_t}{\sqrt{1-\bar{\alpha}_t}} \epsilon_\theta\left(\mathbf{z}_t, t, \mathbf{c}\right)\right), 
\label{eq:ddpm_mean}\vspace{-0.5em}
\end{equation}

\begin{equation}
p_\theta\left(\mathbf{z}_{t-1} \mid \mathbf{z}_t,\mathbf{c}\right)=\mathcal{N}\left(\mu_\theta\left(\mathbf{z}_t, t, \mathbf{c}\right), \Sigma_t\right),
\label{eq:transition}
\end{equation}
where $\alpha_t$, $\beta_t$ are pre-defined constants for timestep dependant denoising, $\Sigma_t$ is covariance matrix of denoising transition, and $\bar{\alpha}_t:=\prod_{s=1}^t \alpha_s$. Lastly, the decoder $\mathcal{D}$ transforms the denoised latent sample $\mathbf{z}_0$ into $\mathbf{x}$ (i.e., pixel space). In the following sections, we treat $\mathbf{z}_0$ as final image, omitting the decoder process for brevity. 

\section{\metabbr}
\begin{figure*}[t]
  \centering
   \includegraphics[width=1.0\linewidth]{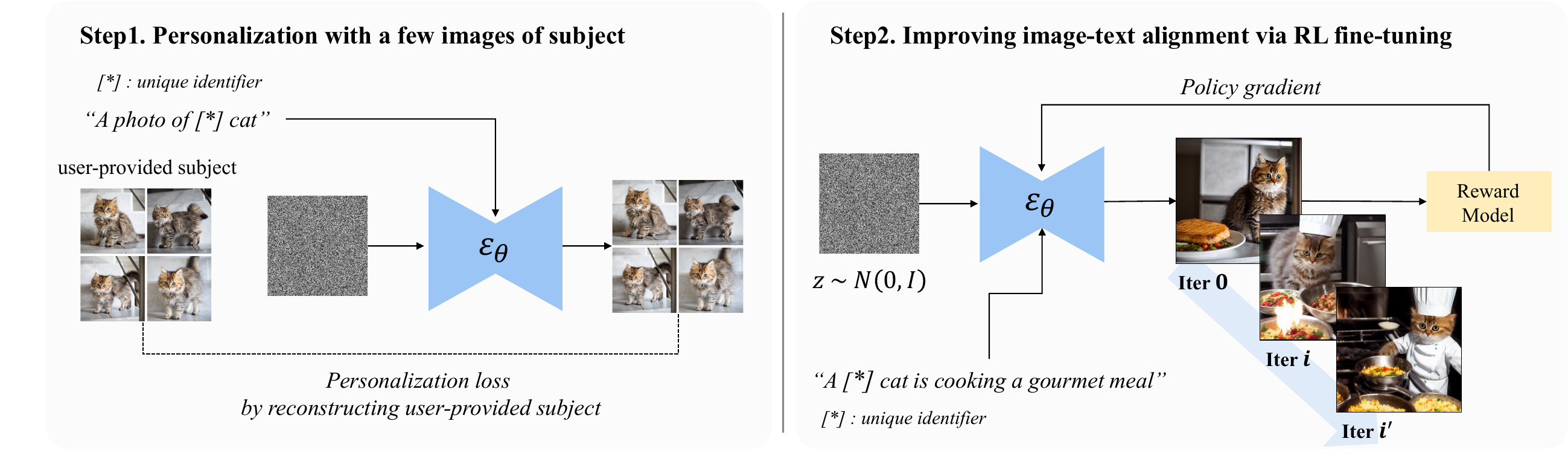}
   \vspace{-1em}
   \caption{\textbf{An overview of \metabbr}. 
   Given a user-specific subject (e.g., a cat) from a few input images, our method enables the personalized text-to-image model to generate new images of that subject with various contexts (e.g., ``a cat is cooking a gourmet meal''). Our method consists of two main steps: (left) Personalization with a few images of subject, where a pre-trained text-to-image model is fine-tuned with a unique identifier and (right) RL fine-tuning for improving image-text alignment, where we further fine-tune the personalized model to maximize the reward that quantifies image-text alignment.}
   \label{fig:overview}
   \vspace{-1em}
   
\end{figure*}

In this section, we describe our approach for enhancing image-text alignment in personalized text-to-image models. Built upon DreamBooth~\cite{dreambooth}, we first personalize a pre-trained text-to-image model with a class noun and a unique identifier (Section~\ref{sec:supervised}).
Next, we fine-tune this personalized model using policy gradient methods~\cite{dpok,ddpo} to maximize a reward that quantifies image-text alignment (Section~\ref{sec:rlbased}).

\subsection{Personalizing Text-to-Image Models}\label{sec:supervised}
\paragraph{Prompts with Unique Identifiers for Personalization.}
To generate new images of a specific subject, given only a few reference images, we leverage DreamBooth~\cite{dreambooth}. DreamBooth fine-tunes a text-to-image diffusion model by associating each user-provided subject with a unique identifier. Formally, we fine-tune the diffusion model by minimizing the following loss:
\begin{align*}
\mathcal{L}_{P} :=\mathbb{E}_{\mathbf{z}, \mathbf{z}^{pr}, \mathbf{c}, \mathbf{c}^{pr}, \epsilon, \epsilon', t, t'}\Big[ &
    \left\|\epsilon-\epsilon_\theta\left(\mathbf{z}_t, t, \mathbf{c}\right)\right\|_2^2 \\ 
    &+ \left\|\epsilon'-\epsilon_\theta\left(\mathbf{z}^{pr}_{t'}, t', \mathbf{c}^{pr}\right)\right\|_2^2 \Big],
\end{align*}
where $\mathbf{c}$ is a prompt with a unique identifier (i.e., ``a [identifier] [class noun]''), $\mathbf{c}^{pr}$ is a prompt consisting of only a class noun (i.e., ``a [class noun]''), $\mathbf{z}_t$ is a noised latent representation of the specific subject image at timestep $t$, and $\mathbf{z}^{pr}_{t'}$ is a noised latent representation of the image representing the class to which a specific subject belongs at timestep $t'$.
The first term represents personalization loss, where the model learns a given subject with a unique identifier (i.e., ``[identifier]''), and the second term is prior-preservation loss, which is used to supervise the model with its own generated images, to retain its visual prior. This fine-tuning process enables the model to generate new images of a subject in different contexts while maintaining its visual prior, which is crucial in performance. 

\paragraph{Detailed Descriptions for Rare Subjects.}
In many cases, when fine-tuning a text-to-image model using a unique identifier with a class noun (i.e., ``a [identifier] [class noun]''), it leads to the generation of perceptually similar personalized images. However, we found that the model often struggles to learn a rare subject with unseen visual attributes. 
For example, when personalizing text-to-image models with the prompt ``a [identifier] plushie'', the model often fails to generate Phryge (the Paris 2024 Olympic mascot) as a personalized output (see Figure~\ref{fig:finegrained} for supporting experimental results). This highlights the challenges posed by rare subjects with unseen visual attributes.
We assume that using a class noun alone may not be sufficient to guide the text-to-image model to learn the distinctive characteristics of such a rare subject. Therefore, we add a detailed description of the subject's attribute to the prompt as the format of ``a [identifier] [description] [class noun]''. We observe that such a simple technique enables text-to-image models to capture the visual characteristics more concretely, resulting in improved personalization capabilities.

\begin{figure*}[t]
  \centering
   \includegraphics[width=0.9\linewidth]{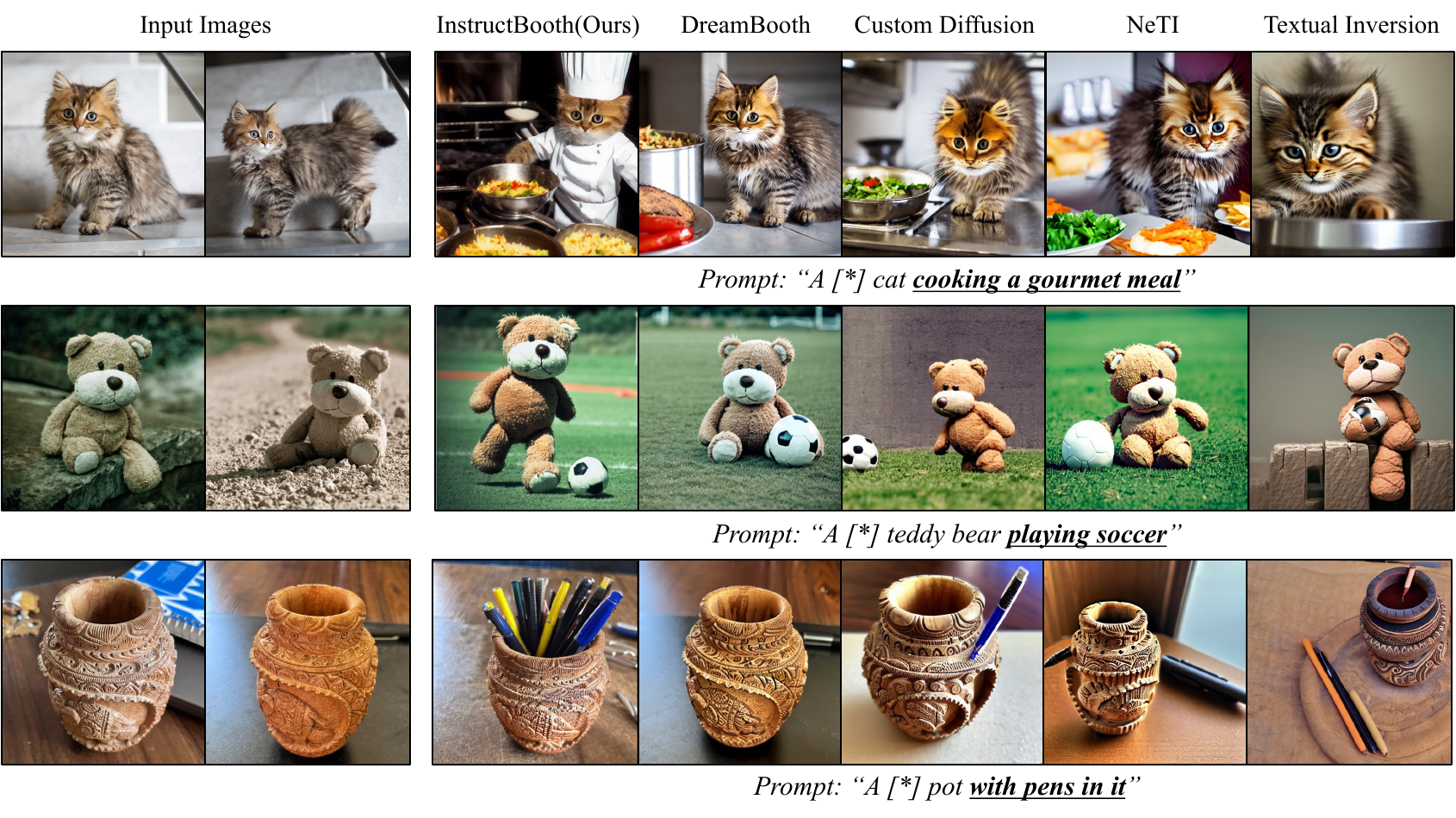}
   \vspace{-1em}
   \caption{Qualitative comparison against DreamBooth, Custom Diffusion, NeTI, and Textual Inversion. Given a few images of a unique subject (e.g., cat, teddy bear, and pot) and a text prompt, models are required to generate personalized images that align with the prompt. [*] denotes a unique identifier. Please see the Appendix~\ref{sup:samples} for more diverse samples.}
   \label{fig:result_comparison_seen} \vspace{-0.3em}
\end{figure*}

\subsection{RL Fine-tuning for Improving Alignment}\label{sec:rlbased}
Fine-tuning a text-to-image model with a unique identifier enables the model to generate new images that are perceptually similar to a user-provided subject. However, such a personalized model often exhibits low text fidelity, especially in terms of contextual diversity, such as subjects' poses, articulations, or interactions with other objects. To address this issue, we propose fine-tuning the personalized model using reinforcement learning (RL).
During the RL fine-tuning step, the personalized model is trained to maximize a reward function reflecting image-text alignment. This training helps the model generate subject images that are closely aligned with the provided prompts.

Formally, similar to existing approaches~\cite{ddpo,dpok}, we formulate the denoising process as multi-step MDP and treat the $p_\theta\left(\mathbf{z}_{t-1} | \mathbf{z}_t,\mathbf{c}\right)$ (described in Eq.~\ref{eq:ddpm_mean},~\ref{eq:transition}) as a policy. 
Based on this formulation, we fine-tune the personalized text-to-image model to maximize the expected reward $r(\mathbf{z}_0, \mathbf{c})$ of the generated images given the prompt distribution $p(\mathbf{c})$:
\begin{equation}
\max_{\theta}~~\mathbb{E}_{p(\mathbf{c}),~p_\theta(\mathbf{z}_0|\mathbf{c})}\left[r\left(\mathbf{z}_0, \mathbf{c}\right)\right],
\label{eq:RL}
\end{equation}
where $p_\theta(\cdot|\mathbf{c})$ is the sample distribution for the final denoised image $\mathbf{z}_0$. The gradient of the RL objective in the Eq.~\ref{eq:RL} can be rewritten as follows: 
\begin{equation}
\mathbb{E}_{p(\mathbf{c}),p_\theta(\mathbf{z}_0|\mathbf{c})}\left[\sum_{t=1}^T \nabla_\theta \log p_\theta\left(\mathbf{z}_{t-1}| \mathbf{z}_t, \mathbf{c}\right)r\left(\mathbf{z}_0, \mathbf{c}\right)\right],
    \label{eq:diffusion_policy}
\end{equation}
where the proof is in~\citet{dpok}. Given the gradient of the RL objective, we update the model parameters $\theta$ to maximize the expected reward.

To measure the alignment of generated images with text prompts, we use ImageReward~\cite{imagereward}, an open-source reward model trained on a large human feedback dataset. \citet{imagereward} demonstrated ImageReward has a better correlation with human judgments compared to other scoring functions, such as CLIP~\cite{clip} and BLIP~\cite{blip}.

\paragraph{Text Prompts for RL Fine-tuning.} 
For RL fine-tuning, we require training text prompts. 
One of the challenges we encounter when dealing with personalized models is the generation of images depicting personalized subjects in various poses. To address this challenge, we propose a templated approach that combines a class noun with a unique identifier along with the phrase describing a specific pose or activity. For example, we use prompts like ``A [identifier] cat is cooking a gourmet meal'' in Figure~\ref{fig:overview}. However, we find that only utilizing prompts with unique identifiers can result in slow RL fine-tuning, as the overfitted model merely generates good samples to get high reward signals. To mitigate this issue, we also utilize prompts without unique identifiers. In summary, we employ both types of prompts during RL fine-tuning: (i) ``a [identifier] [class noun] [activity descriptor]'' and (ii) ``a [class noun] [activity descriptor]''.

\section{Experiments}
\begin{figure*}[ht]
  \centering
   \includegraphics[width=0.95\linewidth]{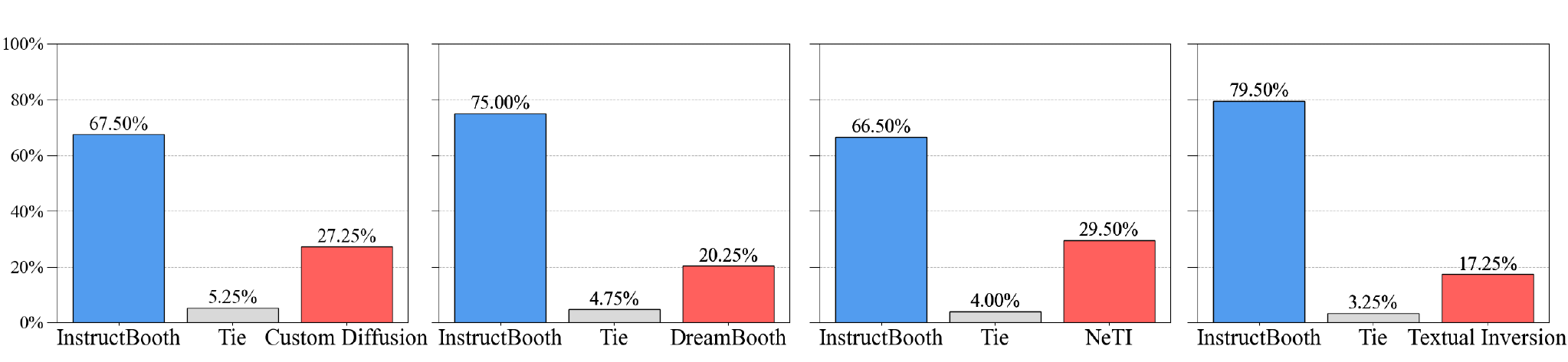}
   \vspace{-1em}
   \caption{Human evaluation results between \metabbr and baselines. Given two images generated by each model, we ask human raters to indicate which is better in overall quality. The results show the preference rates aggregated via majority voting over seven independent human raters.}
   \label{fig:human_overall} \vspace{-0.4em}
\end{figure*}

\subsection{Setup} \label{subsection_setup}
\paragraph{Dataset.}
To evaluate the performance of personalized text-to-image models, we employ the DreamBench dataset~\cite{dreambooth}, which consists of 30 unique subjects including various objects, live subjects, and pets. We evaluate the models using 25 diverse prompts, encompassing recontextualization, property modification, and accessorization.

However, text prompts in DreamBench are not designed to evaluate the model's ability to generate images with actions (e.g., ``a [identifier] [class noun] is cooking a gourmet meal'') or interactions with other objects (e.g., ``a [identifier] [class noun] with popcorn in it''). To address this, we introduce a new dataset comprising eight subjects (teddy bear, cat, dog, monster toy, wooden pot, cup, motorbike, and bike) and five phrases describing activities (e.g., ``playing soccer'' or ``cooking'') or interactions (e.g., ``inserting a pen''). Specifically, we use images of subjects sampled from prior works~\cite{dreambooth, multiconcept} and generate a total of 40 prompts by combining subjects and phrases. We provide details in Appendix~\ref{sup:dataset}. 

\paragraph{Evaluation Metrics.}\label{sec:evaluation}
To assess the quality of text-to-image personalization models, we evaluate them based on two key aspects:
(i) text fidelity, which measures the alignment between the text prompt and the generated image, and (ii) subject fidelity, which measures the preservation of subject details in generated images. For text fidelity, we use three metrics: CLIP-T~\cite{clip}, ImageReward~\cite{imagereward}, and PickScore~\cite{kirstain2023pick}. CLIP-T~\cite{clip} measures the cosine similarity between image and text embeddings. ImageReward and PickScore are scoring functions trained on a large human feedback dataset to measure image-text alignment. For subject fidelity, we use DINO score which measures the cosine similarity between the generated and reference images in ViTS/16 DINO~\cite{dino} embedding spaces.

\paragraph{Implementation Details.}
As our baseline text-to-image model, we use Stable Diffusion v1.5 (SD;~\citealt{sd}) for all methods. During the personalization step, we fine-tune the entire U-Net~\cite{unet}, while keeping the other components of SD frozen. For RL fine-tuning, we apply Low-Rank Adaption (LoRA;~\citealt{lora}), which involves training only the low-rank additional weights, to the U-Net. 
We remark that LoRA fine-tuning is crucial in maintaining the personalization ability. More experimental details are provided in Appendix~\ref{sup:implement}.

\subsection{Qualitative Analysis}
\paragraph{Comparison with Baselines.} 
We compare the quality of personalized text-to-image generation against four baselines: DreamBooth~\cite{dreambooth}, Custom diffusion~\cite{multiconcept}, NeTI~\cite{neti}, and Textual Inversion~\cite{textual_inversion}. As shown in Figure~\ref{fig:result_comparison_seen}, DreamBooth produces images of visually similar subjects but often fails to accurately represent the context from the text prompt.
For example, activities like ``cooking'' and ``playing'' are often not reflected in their generated images. 
Custom Diffusion generates images that align with the text prompt, but it exhibits low similarity to the provided subject. 
In contrast to baselines, \metabbr generates images with high text alignment without sacrificing subject similarity. Moreover,
our method effectively produces personalized images that closely match the intended context, including variations in subject poses (e.g., a teddy bear dribbling the ball) and costumes (e.g., a cat wearing like a chef). Please see Appendix~\ref{sup:samples} for more diverse samples.

\begin{figure*}[t]
  \centering
   \includegraphics[width=\linewidth]{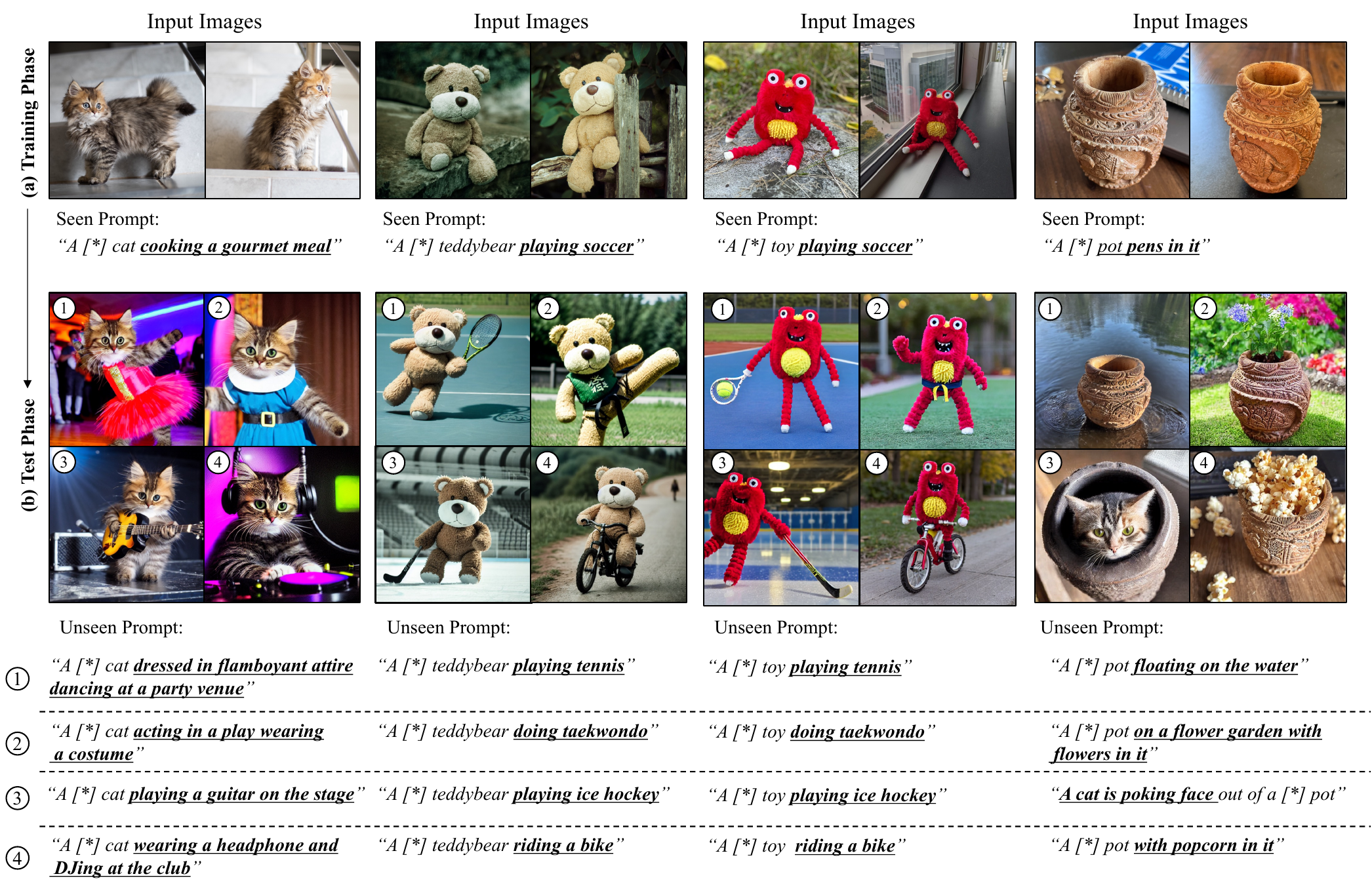}
   \vspace{-2em}
   \caption{Samples generated by \metabbr on unseen text prompts. Our method generates personalized images with high image-text alignment. [*] denotes a unique identifier. We also provide a comparison with other methods in Appendix~\ref{sup:samples}.}
   \label{fig:result_comparison_unseen}
   \vspace{-0.3em}
\end{figure*}

\paragraph{Generalization to Unseen Prompts.}
To understand the generalization ability, we evaluate \metabbr using unseen text prompts. Figure~\ref{fig:result_comparison_unseen} shows image samples from \metabbr on four different subjects with unseen prompts. Our model successfully generates images of user-provided subjects engaged in various unseen activities and costumes, demonstrating the generalization effects of RL fine-tuning. For example, when \metabbr is trained with the prompt ``cooking a gourmet meal'', it can still generate images of the personalized cat engaged in unrelated activities, such as playing a guitar or wearing headphones.

\subsection{Quantitative Analysis}
\paragraph{Human Evaluation.} 
To evaluate the quality of personalized text-to-image generation, we conduct a human study comparing \metabbr with Custom Diffusion~\cite{multiconcept}, DreamBooth~\cite{dreambooth}, NeTI~\cite{neti}, and Textual Inversion~\cite{textual_inversion}.
Using 40 prompts, each consisting of 8 subjects and 5 phrases describing activities (see Section~\ref{subsection_setup} for details), we collect 10 images generated from each prompt, resulting in a total of 400 images for each model. 
First, we ask human raters to provide binary feedback (good/bad)\footnote{Raters were instructed to pass a query if it was difficult to determine. 
In a conservative approach, we considered that what was chosen as ``pass'' was equivalent to being chosen as ``bad''.} in terms of subject fidelity and text fidelity.\footnote{We ask the following questions: (i) Subject fidelity: ``Do the objects in the image closely resemble those in the given (reference) image?'' and (ii) Text fidelity: ``Is the alignment of the image with the text correct?''.} 
Additionally, given two anonymized images (one from \metabbr and one from the baseline) along with a reference image of the subject, human raters indicate 
which image exhibits better overall quality
based on three factors: (i) alignment with the given text (or prompt), (ii) similarity to the object in the reference image, and (iii) naturalness (and quality) of the image. 
Each query is evaluated by seven independent raters using Amazon MTurk and we aggregate the responses via majority voting.
We provide more details of human evaluation in Appendix~\ref{sup:human_eval}.

In Table~\ref{tab:quantitative_score}, the columns labeled ``Human'' indicate the result of our human evaluation, summarizing the binary feedback on subject fidelity and text fidelity. The results show that the performance of baselines is limited due to a trade-off between subject fidelity and text fidelity. In contrast, \metabbr outperforms baselines in terms of text fidelity while exhibiting superior perceptual similarity of subjects. Figure~\ref{fig:human_overall} shows the human preference rating in terms of overall quality. Due to improved text fidelity without compromising on subject fidelity, the images generated by \metabbr are preferred at least twice as much as baselines when considering all relevant factors.

{\setlength{\tabcolsep}{3pt}
\renewcommand{\arraystretch}{1.5}
\begin{table}[t]
    \caption{Comparison with Dreambooth, Custom Diffusion, NeTI and Textual Inversion on our dataset.}
    \vskip 0.15in
    \resizebox{\linewidth}{!}{
    \begin{tabular}{lcccc|cc}
        \toprule
         & \multicolumn{4}{c}{Text Fidelity} & \multicolumn{2}{c}{Subject Fidelity} \\ \hline
        Method & CLIP-T & ImageReward & PickScore & Human & DINO & Human\\
        \hline
        Custom Diffusion & \textbf{0.323} & 1.088 & 0.224 & 91.5\% & 0.535 & 92.8\%\\
        DreamBooth & 0.310 & 0.441 & 0.220 & 81.3\% & \textbf{0.699} & \textbf{97.0\%}\\
        NeTI & 0.311 & 0.480 & 0.218 & 85.5\% & 0.643 & 93.8\%\\
        Textual Inversion & 0.271 & -1.15 & 0.206 & 60.0\% & 0.576 & 80.0\%\\
        \metabbr (Ours) & \textbf{0.323} & \textbf{1.196} & \textbf{0.227} &\textbf{97.3\%} &0.650 &\textbf{97.0\%} \\
        \bottomrule
    \end{tabular}}
    \label{tab:quantitative_score}
    \vspace{-1em}
\end{table} 
}

\paragraph{Evaluation Metrics.} 
We conduct a quantitative evaluation by measuring several metrics introduced in Section~\ref{sec:evaluation}. 
For the evaluation, similar to a human evaluation, we use 40 prompts related to actions to create a total of 400 images, with 10 images for each prompt.
As shown in Table~\ref{tab:quantitative_score}, \metabbr achieves the highest scores in all metrics related to text fidelity. Furthermore, unlike Custom Diffusion which excels in text fidelity but struggles with subject fidelity, our method maintains competitive subject fidelity.
However, in contrast to human evaluation results, our method exhibits a slightly lower score in subject fidelity compared to DreamBooth. 
This difference can be attributed to the limitations of the subject fidelity metric, i.e., DINO, which penalizes changes in pose and primarily considers appearance resemblance.
In our tested prompts that emphasize actions, even if the personalized image is successfully created, it may receive a lower DINO score.
To provide further support for this explanation, we include the images and their corresponding scores in Appendix~\ref{sup:metric}.

{
\setlength{\tabcolsep}{4pt}
\renewcommand{\arraystretch}{1.3}
\begin{table}[t]
    \caption{Comparison with baselines on DreamBench. 
    For baselines, we report both performances reported in \citet{dreambooth} and those obtained through our implementation (denoted as our impl).}
    \vskip 0.15in 
    \resizebox{\linewidth}{!}{
    \begin{tabular}{lcccc}
        \toprule
         & \multicolumn{3}{c}{Text Fidelity} & \multicolumn{1}{c}{Subject Fidelity} \\\hline
        Method & CLIP-T& ImageReward & PickScore & DINO\\
        \hline
        Textual Inversion & 0.255 & N/A & N/A & 0.569\\
        Textual Inversion (our impl) & 0.257 & -1.337 & 0.203 & 0.537\\
        DreamBooth & 0.305 & N/A & N/A & \textbf{0.668}\\
        DreamBooth (our impl) & 0.297 & 0.052 & 0.214 & 0.648\\
        \metabbr (Ours) & \textbf{0.306} & \textbf{0.651} & \textbf{0.217} & 0.661\\
        \bottomrule
    \end{tabular}}
    \label{tab:Dreambench} \vspace{-1em}
\end{table}
}

\paragraph{DreamBench Results.}
In addition to the experiments with our dataset, we also evaluate \metabbr with DreamBench~\cite{dreambooth} dataset to demonstrate that our method can enhance text fidelity in diverse prompt scenarios. Following DreamBench, we generate four images for each subject and each prompt, resulting in a total of 3,000 images for measurement. As for our prompts for RL fine-tuning, we set the two training prompts regarding recontextualization and accessorization which DreamBench primarily deals with. Note that we do not use DreamBench's prompts directly as training prompts, which means that all evaluation is performed with unseen prompts to our method. 
As shown in Table~\ref{tab:Dreambench}, \metabbr achieves the highest scores in text fidelity, while also demonstrating competitive performances in terms of subject fidelity. We provide the generated samples by \metabbr and the baselines in Appendix~\ref{sup:dreambench}.

\subsection{Ablation Study}
\begin{figure}[t]
  \centering
   \includegraphics[width=1.0\linewidth]{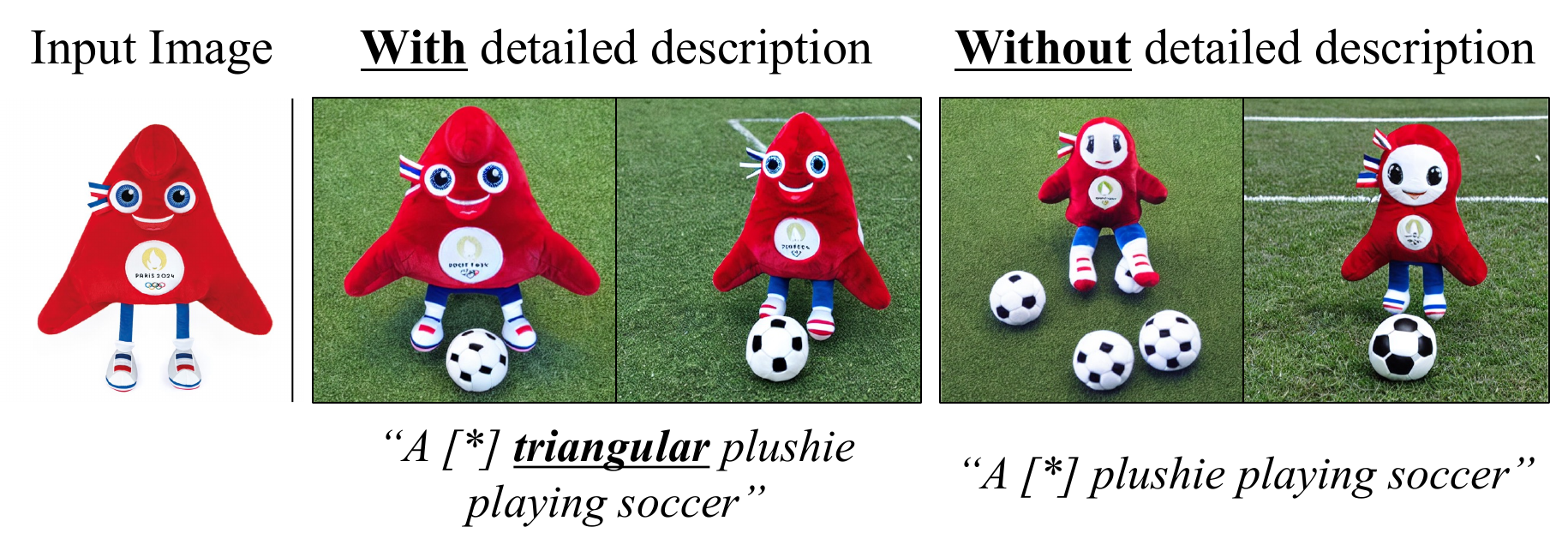}
   \vspace{-2em}
   \caption{Comparison of results generated by models trained with and without prompts including a detailed description (i.e., {\bf triangular}).
   [*] denotes a unique identifier.}
   \label{fig:finegrained}
   \vspace{-1em}
\end{figure}

\paragraph{Effects of Detailed Descriptions in Personalization.}
To demonstrate the importance of detailed description when personalizing text-to-image models with a rare subject, we conduct an ablation study on personalization techniques. 
In this study, we use Phryge (the Paris 2024 Olympic mascot) as our target subject. Since this mascot was not released when our base text-to-image model (i.e., Stable Diffusion v1.5) was trained, it can be considered a truly rare concept for the text-to-image model to learn.
We compare the model trained with a detailed prompt ``a [*] triangular plushie'' and that trained with a standard simple prompt ``a [*] plushie'' using images of Phryge.
As shown in Figure~\ref{fig:finegrained}, the model trained with a standard simple prompt fails to accurately represent the subject's features. Further quantitative analysis of this personalization technique is provided in Appendix~\ref{sup:ablation}.

\begin{figure}[t]
  \centering 
\includegraphics[width=1.0\linewidth]{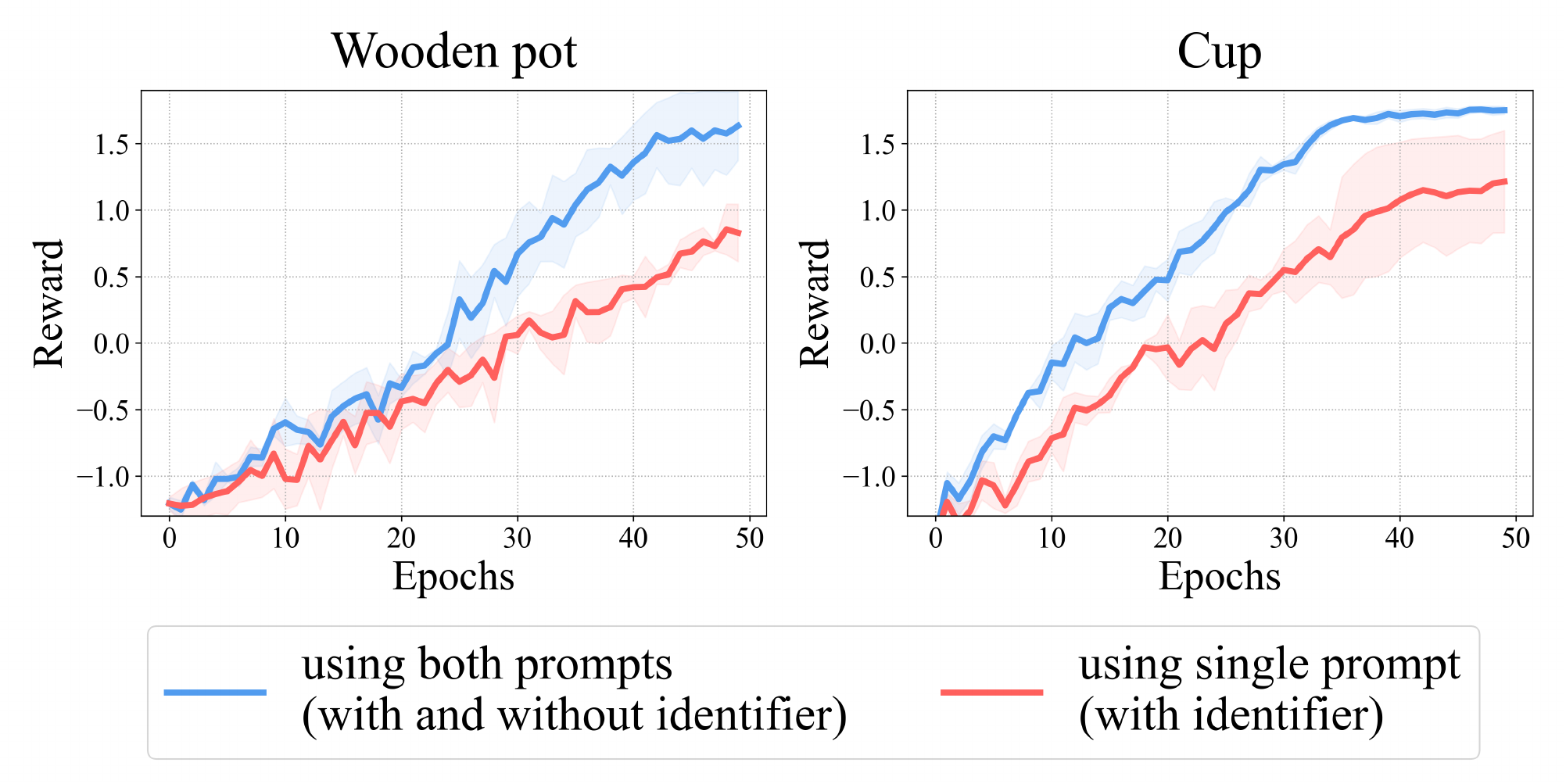}
   \vspace{-2em}
   \caption{Learning curves of the model trained with a single prompt (i.e., ``a [*] [class noun] with pens in it'') and the model trained with both prompts (i.e., including ``a [class noun] with pens in it''). The solid line and shaded regions represent the mean and standard deviation across three runs, respectively.}
   \label{fig:learning_curve}
   \vspace{-1em}
\end{figure}

\paragraph{Text Prompts in RL Fine-tuning.}
In Section~\ref{sec:rlbased}, we introduce an important trick to leverage both text prompts, with and without a unique identifier for RL fine-tuning. 
To verify its effectiveness, we compare two models: 
one trained using a prompt with a unique identifier and another trained using both prompts, with and without a unique identifier. 
In this comparison, we use ``Wooden pot'' and ``Cup'' as the target subject, and set ``with the pens in it'' as the target prompt. 
As shown in Figure~\ref{fig:learning_curve}, the model trained with both prompts exhibits improved sample efficiency, demonstrating that our proposed technique mitigates the challenges faced by overfitted personalized models that often struggle to provide good reward signals during RL fine-tuning.

\section{Conclusion}
In this paper, we propose \metabbr, a new method for improving the image-text alignment of personalized text-to-image models.
We demonstrate that fine-tuning the model with RL in a subsequent stage mitigates overfitting, enabling the model to generate images of specific subjects with contextual diversity in terms of poses and interactions. In qualitative comparison and human evaluation, we show that \metabbr can generate images that are more aligned with human preferences than those of existing models. 
We believe that our approach of subsequent fine-tuning broadens the potential of personalized text-to-image models by allowing the usage of diverse prompts.

\newpage
\section*{Broader Impact}
Personalizing text-to-image models enables users to generate images based on their own chosen subjects, opening up possibilities for user-specific artwork and entertainment. However, there are also concerns about its potential misuse, such as creating deepfakes and infringing on copyrights. Our research aims to improve the performance of personalized text-to-image models by increasing text fidelity. This improvement could amplify both the anticipated benefits and the associated concerns. To mitigate these concerns, it is crucial to advance research in the detection of synthetic data and in the incorporation of watermarking into generative models.


\nocite{langley00}

\bibliography{main}

\begin{thebibliography}{32}
\providecommand{\natexlab}[1]{#1}
\providecommand{\url}[1]{\texttt{#1}}
\expandafter\ifx\csname urlstyle\endcsname\relax
  \providecommand{\doi}[1]{doi: #1}\else
  \providecommand{\doi}{doi: \begingroup \urlstyle{rm}\Url}\fi

\bibitem[Agarwal et~al.(2023)Agarwal, Karanam, Shukla, and Srinivasan]{agarwal2023image}
Agarwal, A., Karanam, S., Shukla, T., and Srinivasan, B.~V.
\newblock An image is worth multiple words: Multi-attribute inversion for constrained text-to-image synthesis.
\newblock \emph{arXiv preprint arXiv:2311.11919}, 2023.

\bibitem[Alaluf et~al.(2023)Alaluf, Richardson, Metzer, and Cohen-Or]{neti}
Alaluf, Y., Richardson, E., Metzer, G., and Cohen-Or, D.
\newblock A neural space-time representation for text-to-image personalization.
\newblock \emph{arXiv preprint arXiv:2305.15391}, 2023.

\bibitem[Black et~al.(2023)Black, Janner, Du, Kostrikov, and Levine]{ddpo}
Black, K., Janner, M., Du, Y., Kostrikov, I., and Levine, S.
\newblock Training diffusion models with reinforcement learning.
\newblock \emph{arXiv preprint arXiv:2305.13301}, 2023.

\bibitem[Caron et~al.(2021)Caron, Touvron, Misra, J{\'e}gou, Mairal, Bojanowski, and Joulin]{dino}
Caron, M., Touvron, H., Misra, I., J{\'e}gou, H., Mairal, J., Bojanowski, P., and Joulin, A.
\newblock Emerging properties in self-supervised vision transformers.
\newblock In \emph{IEEE/CVF international conference on computer vision}, 2021.

\bibitem[Chang et~al.(2023)Chang, Zhang, Barber, Maschinot, Lezama, Jiang, Yang, Murphy, Freeman, Rubinstein, et~al.]{muse}
Chang, H., Zhang, H., Barber, J., Maschinot, A., Lezama, J., Jiang, L., Yang, M.-H., Murphy, K., Freeman, W.~T., Rubinstein, M., et~al.
\newblock Muse: Text-to-image generation via masked generative transformers.
\newblock \emph{arXiv preprint arXiv:2301.00704}, 2023.

\bibitem[Clark et~al.(2023)Clark, Vicol, Swersky, and Fleet]{draft}
Clark, K., Vicol, P., Swersky, K., and Fleet, D.~J.
\newblock Directly fine-tuning diffusion models on differentiable rewards.
\newblock \emph{arXiv preprint arXiv:2309.17400}, 2023.

\bibitem[Fan et~al.(2023)Fan, Watkins, Du, Liu, Ryu, Boutilier, Abbeel, Ghavamzadeh, Lee, and Lee]{dpok}
Fan, Y., Watkins, O., Du, Y., Liu, H., Ryu, M., Boutilier, C., Abbeel, P., Ghavamzadeh, M., Lee, K., and Lee, K.
\newblock Dpok: Reinforcement learning for fine-tuning text-to-image diffusion models.
\newblock In \emph{Advances in neural information processing systems}, 2023.

\bibitem[Gal et~al.(2022)Gal, Alaluf, Atzmon, Patashnik, Bermano, Chechik, and Cohen-Or]{textual_inversion}
Gal, R., Alaluf, Y., Atzmon, Y., Patashnik, O., Bermano, A.~H., Chechik, G., and Cohen-Or, D.
\newblock An image is worth one word: Personalizing text-to-image generation using textual inversion.
\newblock \emph{arXiv preprint arXiv:2208.01618}, 2022.

\bibitem[Han et~al.(2023)Han, Li, Zhang, Milanfar, Metaxas, and Yang]{svdiff}
Han, L., Li, Y., Zhang, H., Milanfar, P., Metaxas, D., and Yang, F.
\newblock Svdiff: Compact parameter space for diffusion fine-tuning.
\newblock \emph{arXiv preprint arXiv:2303.11305}, 2023.

\bibitem[Ho et~al.(2020)Ho, Jain, and Abbeel]{ddpm}
Ho, J., Jain, A., and Abbeel, P.
\newblock Denoising diffusion probabilistic models.
\newblock In \emph{Advances in neural information processing systems}, 2020.

\bibitem[Hu et~al.(2021)Hu, Shen, Wallis, Allen-Zhu, Li, Wang, Wang, and Chen]{lora}
Hu, E.~J., Shen, Y., Wallis, P., Allen-Zhu, Z., Li, Y., Wang, S., Wang, L., and Chen, W.
\newblock Lora: Low-rank adaptation of large language models.
\newblock \emph{arXiv preprint arXiv:2106.09685}, 2021.

\bibitem[Kirstain et~al.(2023)Kirstain, Polyak, Singer, Matiana, Penna, and Levy]{kirstain2023pick}
Kirstain, Y., Polyak, A., Singer, U., Matiana, S., Penna, J., and Levy, O.
\newblock Pick-a-pic: An open dataset of user preferences for text-to-image generation.
\newblock In \emph{Advances in neural information processing systems}, 2023.

\bibitem[Kumari et~al.(2023)Kumari, Zhang, Zhang, Shechtman, and Zhu]{multiconcept}
Kumari, N., Zhang, B., Zhang, R., Shechtman, E., and Zhu, J.-Y.
\newblock Multi-concept customization of text-to-image diffusion.
\newblock In \emph{IEEE/CVF Conference on Computer Vision and Pattern Recognition}, 2023.

\bibitem[Langley(2000)]{langley00}
Langley, P.
\newblock Crafting papers on machine learning.
\newblock In Langley, P. (ed.), \emph{Proceedings of the 17th International Conference on Machine Learning (ICML 2000)}, pp.\  1207--1216, Stanford, CA, 2000. Morgan Kaufmann.

\bibitem[Lee et~al.(2023)Lee, Liu, Ryu, Watkins, Du, Boutilier, Abbeel, Ghavamzadeh, and Gu]{lee2023aligning}
Lee, K., Liu, H., Ryu, M., Watkins, O., Du, Y., Boutilier, C., Abbeel, P., Ghavamzadeh, M., and Gu, S.~S.
\newblock Aligning text-to-image models using human feedback.
\newblock \emph{arXiv preprint arXiv:2302.12192}, 2023.

\bibitem[Li et~al.(2022)Li, Li, Xiong, and Hoi]{blip}
Li, J., Li, D., Xiong, C., and Hoi, S.
\newblock Blip: Bootstrapping language-image pre-training for unified vision-language understanding and generation.
\newblock In \emph{International Conference on Machine Learning}, 2022.

\bibitem[Liu et~al.(2023)Liu, Feng, Zhu, Zhang, Zheng, Liu, Zhao, Zhou, and Cao]{cones}
Liu, Z., Feng, R., Zhu, K., Zhang, Y., Zheng, K., Liu, Y., Zhao, D., Zhou, J., and Cao, Y.
\newblock Cones: Concept neurons in diffusion models for customized generation.
\newblock \emph{arXiv preprint arXiv:2303.05125}, 2023.

\bibitem[Loshchilov \& Hutter(2019)Loshchilov and Hutter]{loshchilov2017decoupled}
Loshchilov, I. and Hutter, F.
\newblock Decoupled weight decay regularization.
\newblock \emph{ICLR}, 2019.

\bibitem[Prabhudesai et~al.(2023)Prabhudesai, Goyal, Pathak, and Fragkiadaki]{alignprop}
Prabhudesai, M., Goyal, A., Pathak, D., and Fragkiadaki, K.
\newblock Aligning text-to-image diffusion models with reward backpropagation.
\newblock \emph{arXiv preprint arXiv:2310.03739}, 2023.

\bibitem[Qiu et~al.(2023)Qiu, Liu, Feng, Xue, Feng, Liu, Zhang, Weller, and Sch{\"o}lkopf]{orthogonal}
Qiu, Z., Liu, W., Feng, H., Xue, Y., Feng, Y., Liu, Z., Zhang, D., Weller, A., and Sch{\"o}lkopf, B.
\newblock Controlling text-to-image diffusion by orthogonal finetuning.
\newblock In \emph{Thirty-seventh Conference on Neural Information Processing Systems}, 2023.
\newblock URL \url{https://openreview.net/forum?id=K30wTdIIYc}.

\bibitem[Radford et~al.(2021)Radford, Kim, Hallacy, Ramesh, Goh, Agarwal, Sastry, Askell, Mishkin, Clark, et~al.]{clip}
Radford, A., Kim, J.~W., Hallacy, C., Ramesh, A., Goh, G., Agarwal, S., Sastry, G., Askell, A., Mishkin, P., Clark, J., et~al.
\newblock Learning transferable visual models from natural language supervision.
\newblock In \emph{International conference on machine learning}, 2021.

\bibitem[Ramesh et~al.(2022)Ramesh, Dhariwal, Nichol, Chu, and Chen]{dalle2}
Ramesh, A., Dhariwal, P., Nichol, A., Chu, C., and Chen, M.
\newblock Hierarchical text-conditional image generation with clip latents.
\newblock \emph{arXiv preprint arXiv:2204.06125}, 2022.

\bibitem[Rombach et~al.(2022)Rombach, Blattmann, Lorenz, Esser, and Ommer]{sd}
Rombach, R., Blattmann, A., Lorenz, D., Esser, P., and Ommer, B.
\newblock High-resolution image synthesis with latent diffusion models.
\newblock In \emph{IEEE/CVF Conference on Computer Vision and Pattern Recognition}, 2022.

\bibitem[Ronneberger et~al.(2015)Ronneberger, Fischer, and Brox]{unet}
Ronneberger, O., Fischer, P., and Brox, T.
\newblock U-net: Convolutional networks for biomedical image segmentation.
\newblock In \emph{Medical Image Computing and Computer-Assisted Intervention}, 2015.

\bibitem[Ruiz et~al.(2023)Ruiz, Li, Jampani, Pritch, Rubinstein, and Aberman]{dreambooth}
Ruiz, N., Li, Y., Jampani, V., Pritch, Y., Rubinstein, M., and Aberman, K.
\newblock Dreambooth: Fine tuning text-to-image diffusion models for subject-driven generation.
\newblock In \emph{IEEE/CVF Conference on Computer Vision and Pattern Recognition}, 2023.

\bibitem[Saharia et~al.(2022)Saharia, Chan, Saxena, Li, Whang, Denton, Ghasemipour, Gontijo~Lopes, Karagol~Ayan, Salimans, et~al.]{imagen}
Saharia, C., Chan, W., Saxena, S., Li, L., Whang, J., Denton, E.~L., Ghasemipour, K., Gontijo~Lopes, R., Karagol~Ayan, B., Salimans, T., et~al.
\newblock Photorealistic text-to-image diffusion models with deep language understanding.
\newblock In \emph{Advances in Neural Information Processing Systems}, 2022.

\bibitem[Tewel et~al.(2023)Tewel, Gal, Chechik, and Atzmon]{key_locked}
Tewel, Y., Gal, R., Chechik, G., and Atzmon, Y.
\newblock Key-locked rank one editing for text-to-image personalization.
\newblock In \emph{ACM SIGGRAPH 2023 Conference Proceedings}, 2023.

\bibitem[Voynov et~al.(2023)Voynov, Chu, Cohen-Or, and Aberman]{P+}
Voynov, A., Chu, Q., Cohen-Or, D., and Aberman, K.
\newblock $ p+ $: Extended textual conditioning in text-to-image generation.
\newblock \emph{arXiv preprint arXiv:2303.09522}, 2023.

\bibitem[Wu et~al.(2023)Wu, Sun, Zhu, Zhao, and Li]{hps}
Wu, X., Sun, K., Zhu, F., Zhao, R., and Li, H.
\newblock Human preference score: Better aligning text-to-image models with human preference.
\newblock In \emph{IEEE/CVF International Conference on Computer Vision}, 2023.

\bibitem[Xu et~al.(2023)Xu, Liu, Wu, Tong, Li, Ding, Tang, and Dong]{imagereward}
Xu, J., Liu, X., Wu, Y., Tong, Y., Li, Q., Ding, M., Tang, J., and Dong, Y.
\newblock Imagereward: Learning and evaluating human preferences for text-to-image generation.
\newblock In \emph{Advances in neural information processing systems}, 2023.

\bibitem[Yu et~al.(2022)Yu, Xu, Koh, Luong, Baid, Wang, Vasudevan, Ku, Yang, Ayan, et~al.]{parti}
Yu, J., Xu, Y., Koh, J.~Y., Luong, T., Baid, G., Wang, Z., Vasudevan, V., Ku, A., Yang, Y., Ayan, B.~K., et~al.
\newblock Scaling autoregressive models for content-rich text-to-image generation.
\newblock \emph{arXiv preprint arXiv:2206.10789}, 2022.

\bibitem[Zhang et~al.(2023)Zhang, Dong, Tang, Huang, Huang, Ma, Lee, Deussen, and Xu]{prospect}
Zhang, Y., Dong, W., Tang, F., Huang, N., Huang, H., Ma, C., Lee, T.-Y., Deussen, O., and Xu, C.
\newblock Prospect: Expanded conditioning for the personalization of attribute-aware image generation.
\newblock \emph{arXiv preprint arXiv:2305.16225}, 2023.

\end{thebibliography}
\bibliographystyle{icml2024}

\newpage
\appendix
\onecolumn

\section{Dataset Details}~\label{sup:dataset}
\begin{figure*}[ht]
    \centering
    \includegraphics[width=1\linewidth]{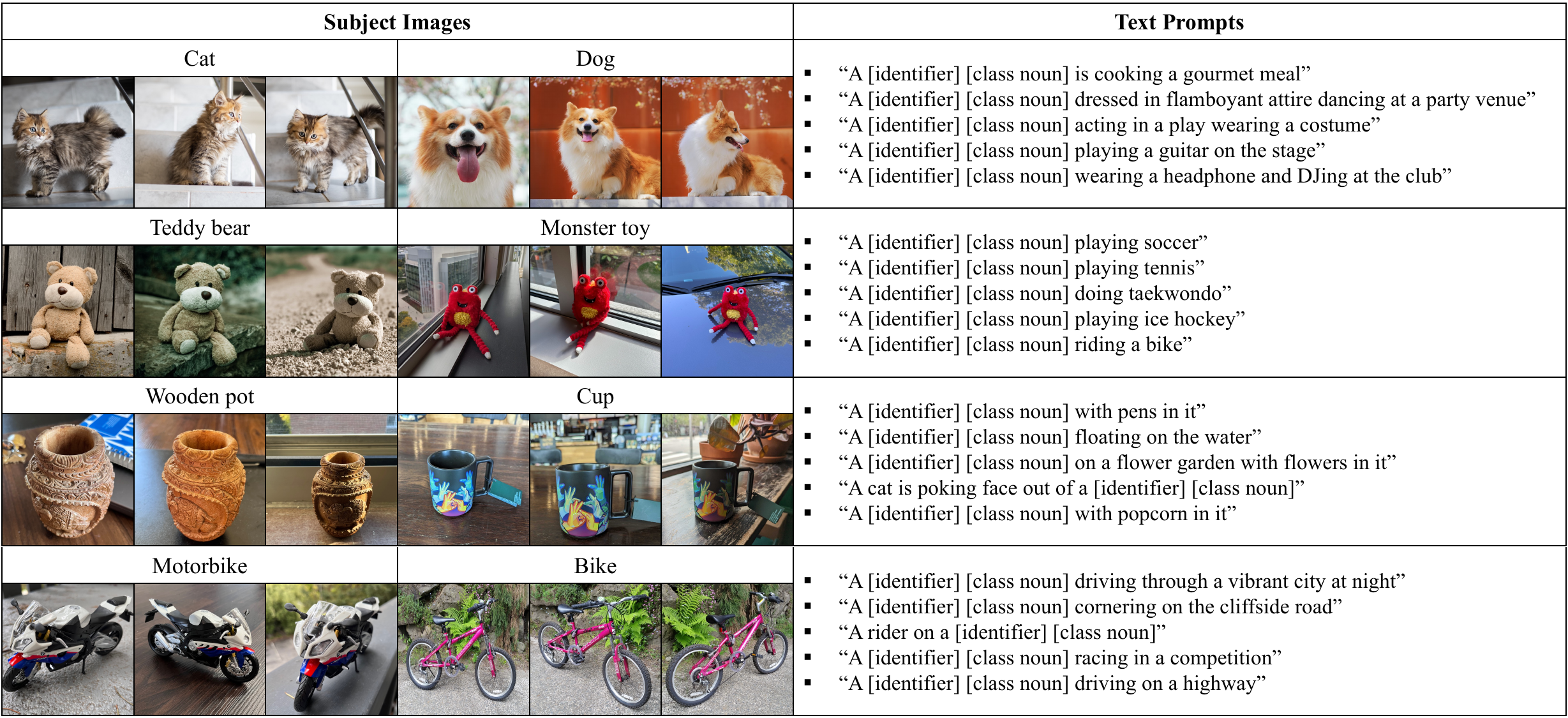}
    \caption{Our newly created evaluation dataset (for text-to-image personalization methods) that contains phrases focusing on actions or interactions with other objects. Our dataset consists of eight subjects (i.e., cat, dog, teddy bear, monster toy, wooden pot, cup, motorbike, and bike) and five corresponding phrases describing activities (e.g., ``playing soccer,'' ``riding a bike,'' and ``cooking'') or interactions (e.g., ``floating on the water'' and ``with popcorn in it'').}
    \label{fig:dataset_examples_appendix}
\end{figure*}

As we explained in the main paper, the current DreamBench~\cite{dreambooth} dataset is not well-suited to evaluate the model's ability to generate images regarding actions or interactions with other objects. Thus, as shown in Figure~\ref{fig:dataset_examples_appendix}, we introduce a new dataset, which consists of eight subjects (i.e., cat, dog, teddy bear, monster toy, wooden pot, cup, motorbike, and bike) and five corresponding phrases describing activities (e.g., ``playing soccer,'' ``riding a bike,'' and ``cooking'') and interactions (e.g., ``floating on the water'' and ``with popcorn in it''). Overall, we have $40$ ($=8\times5$) object-phrase pairs. Note that we use images of a dog and a monster toy from \citet{dreambooth} and images of a cat, a teddy bear, a wooden pot, a cup, a motorbike, and a bike from \citet{multiconcept}. Further, we group two similar subjects (e.g., a cat and a dog) and use the same five phrases for each group. 

\section{Implementation Details}~\label{sup:implement}
As for personalization step, our fine-tuning pipeline is built upon the publicly available repository.\footnote{\href{https://github.com/huggingface/diffusers/tree/main/examples/dreambooth}{https://github.com/huggingface/diffusers/tree/main/examples/dreambooth}} We generate 200 images representing the class to which input subject belongs for prior-preservation loss. We set the learning rate to $2\times10^{-5}$ and batch size to $2$ (one is an input subject image for personalization loss, and the other is a class image for prior-preservation loss). We update the model using AdamW~\cite{loshchilov2017decoupled}, where $\beta_{1}=0.9$, $\beta_{2}=0.99$ and weight decay $0.01$.

As for RL fine-tuning step, our implementation is based on DDPO~\cite{ddpo}. We generate $16$ images at each epoch using a personalized diffusion model with an inference step of $50$ and a guidance scale of $7.5$. Among the denoising trajectories of $16$ generated images, we randomly use $8$ samples as the training batch at each gradient step. For policy gradient-based optimization, we use importance sampling with a clip range of $1\times10^{-4}$. We set the learning rate to $2\times10^{-5}$ and update the model using AdamW with the same settings in the personalization step.

\newpage
\section{Analysis of Subject Fidelity Metric}~\label{sup:metric}
\begin{figure*}[h]
    \centering
    \includegraphics[width=1.0\linewidth]{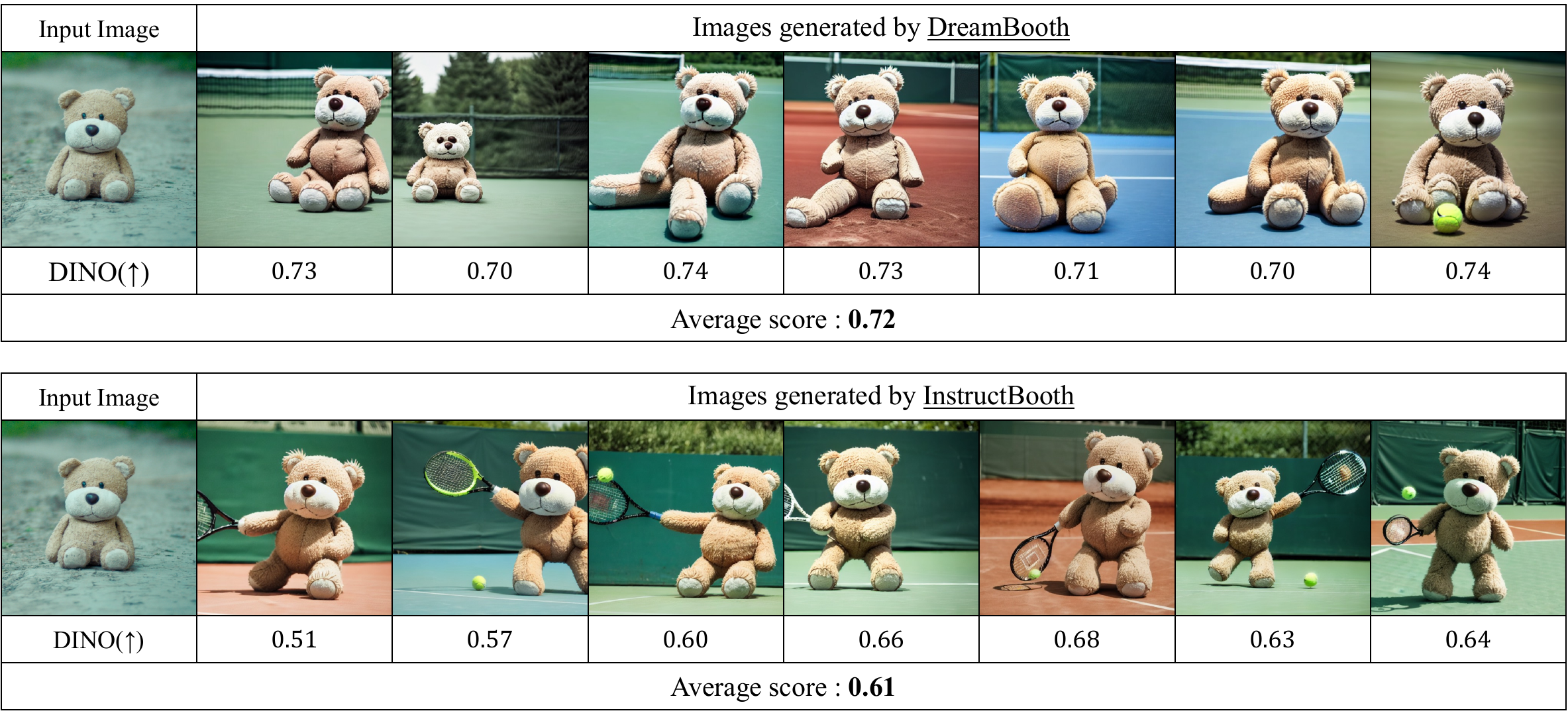}
    \caption{Examples of generated personalized images by DreamBooth~\cite{dreambooth} and InstructBooth (ours), using input image (left). We report subject fidelity scores (DINO~\cite{dino}) for each generated image. Note that $\uparrow$ indicates that the higher number is the better.}
    \label{fig:subject_metric_limit}
\end{figure*}

In the main paper, we argue that subject fidelity metric (i.e., DINO~\cite{dino}) might not be a perfect metric to evaluate the quality of generated personalized images. In our analysis, this metric gives the best score for copied-and-pasted subjects, giving lower scores for the same subjects of different poses or actions. 
To demonstrate it, we present the generated results of DreamBooth~\cite{dreambooth} and InstructBooth along with the corresponding scores. For this analysis, we use ``A [identifier] teddybear playing tennis'' as a text prompt and collect $7$ images of each method. Note that we intentionally collect DreamBooth's results containing unnatural visual elements to show that changes in pose incur a greater penalty than failure factors of personalization. 
As shown in Figure~\ref{fig:subject_metric_limit}, even though examples from DreamBooth are perceptually \textit{less} similar to the reference subject (e.g., a teddy bear with three legs), their DINO scores are clearly shown better than ours, where we generate perceptually \textit{more} similar subjects with different actions. This may indicate that such subject fidelity metric may not be useful to accurately quantify perceptual similarities of subjects with different poses.

\section{Additional Generated Examples}~\label{sup:samples}
\begin{figure*}
    \centering
    \includegraphics[width=0.95\linewidth]{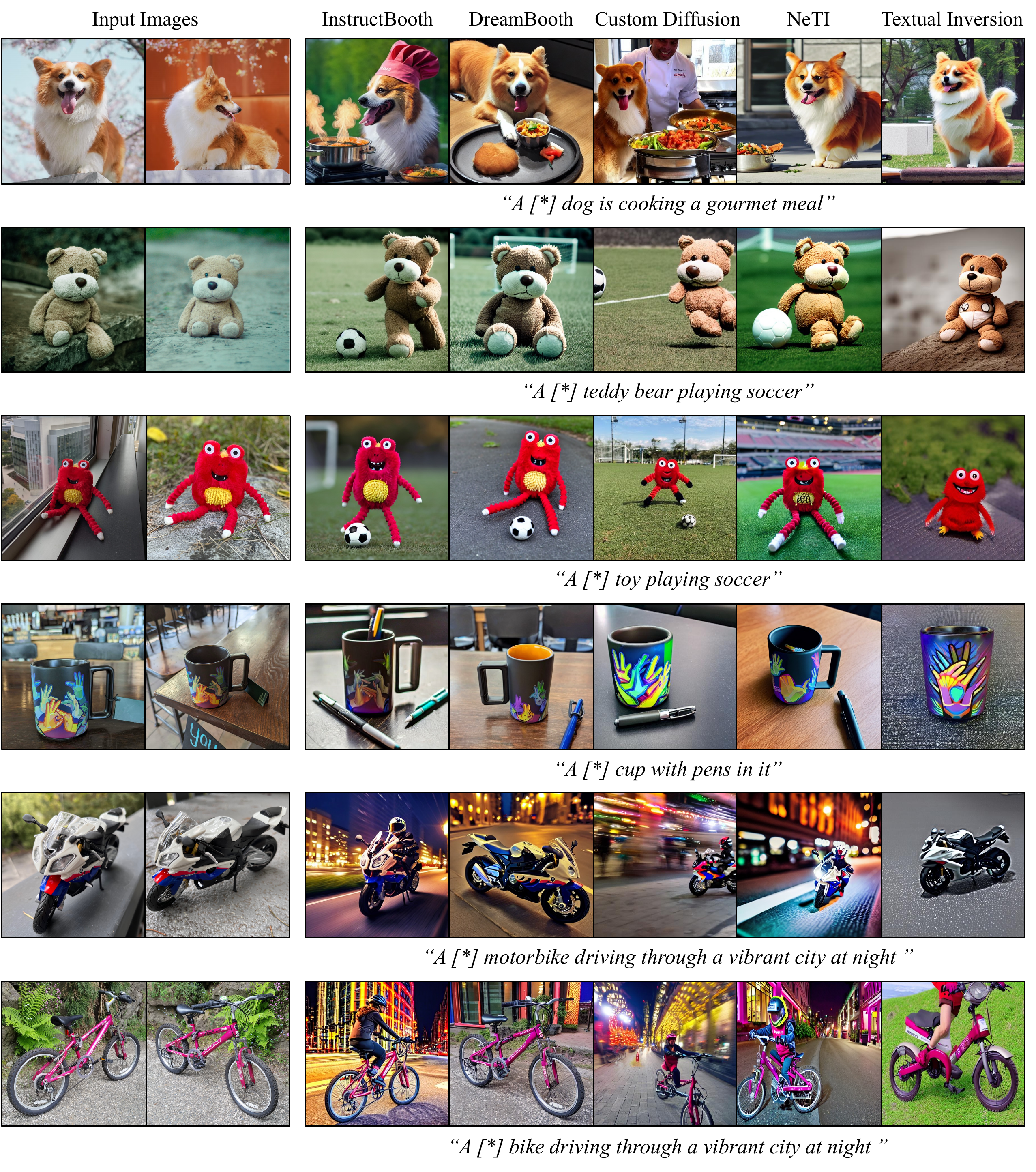}
    \caption{Additional examples generated by InstructBooth, DreamBooth~\cite{dreambooth}, Custom Diffusion~\cite{multiconcept}, NeTI~\cite{neti}, and Textual Inversion~\cite{textual_inversion}. Note that the prompts used to generate these samples are used in the RL fine-tuning step of InstructBooth (i.e., \textit{seen} prompt).}
    \label{fig:seen_examples_sup}
\end{figure*}

\begin{figure*}
    \centering
    \includegraphics[width=0.95\linewidth]{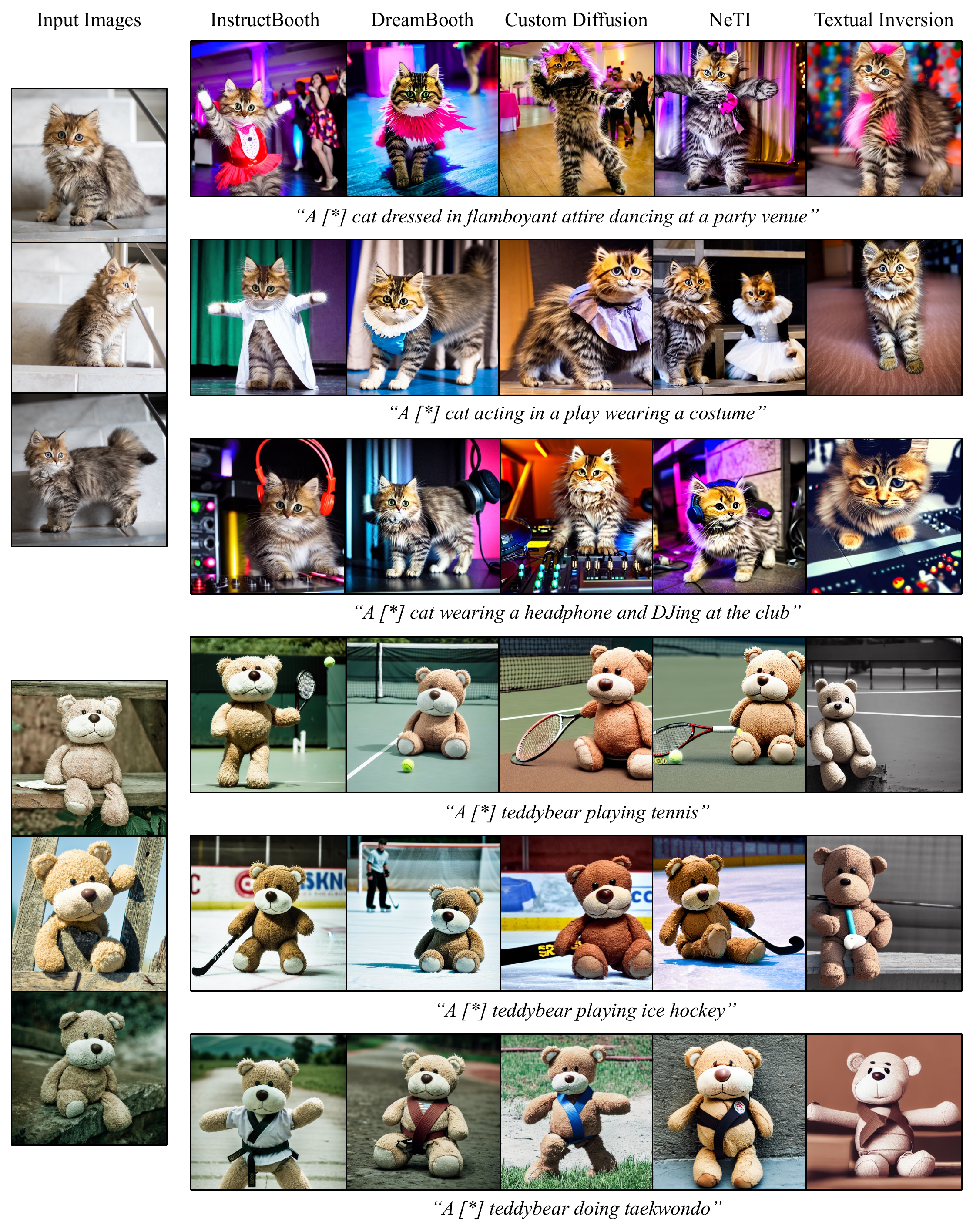}
    \caption{Additional examples generated by InstructBooth, DreamBooth~\cite{dreambooth}, Custom Diffusion~\cite{multiconcept}, NeTI~\cite{neti}, and Textual Inversion~\cite{textual_inversion}. Note that the prompts used to generate these samples are not used in the RL fine-tuning step of InstructBooth (i.e., \textit{unseen} prompt).}
    \label{fig:unseen_examples_sup_1}
\end{figure*}

\begin{figure*}
    \centering
    \includegraphics[width=0.95\linewidth]{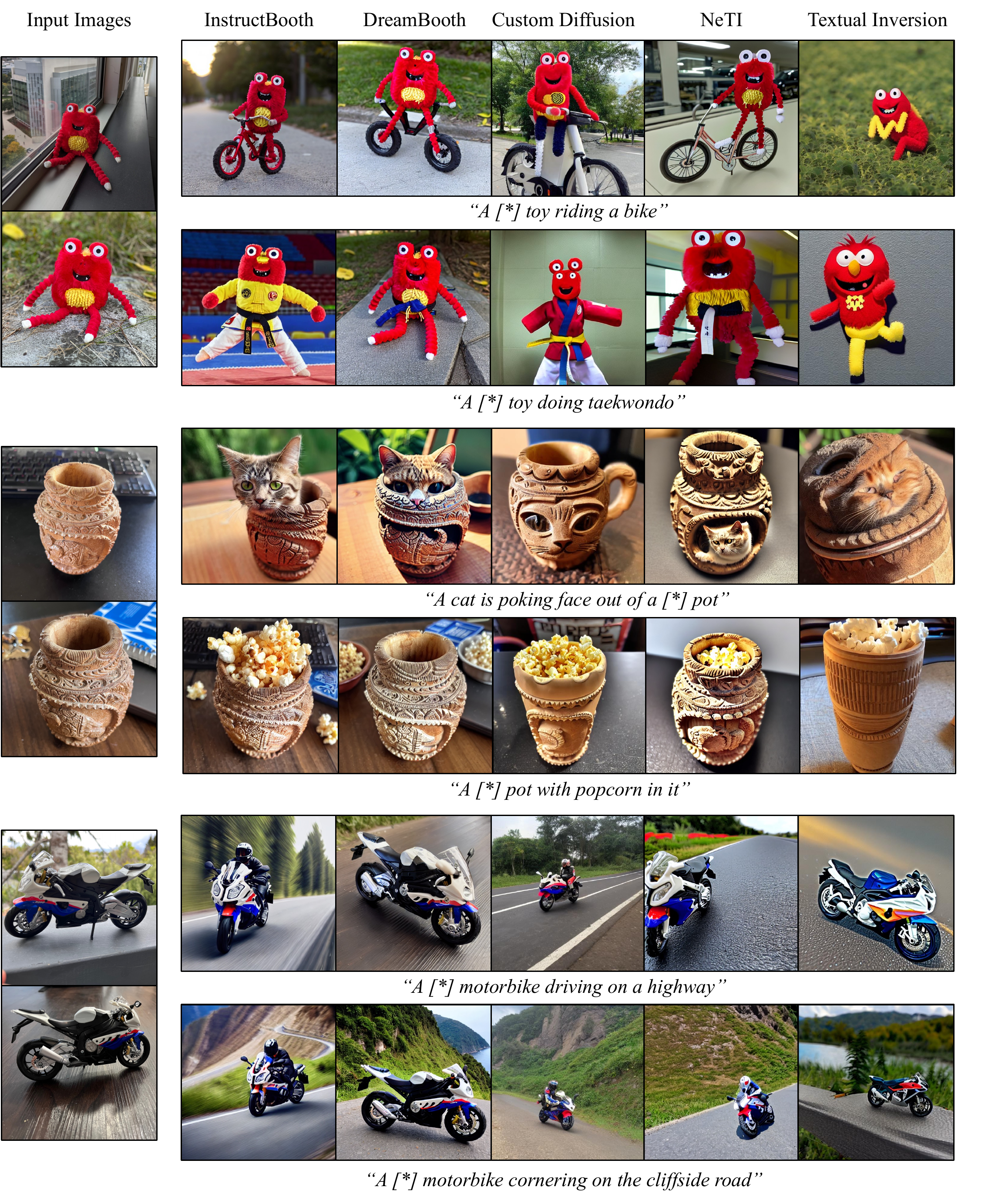}
    \caption{Additional examples generated by InstructBooth, DreamBooth~\cite{dreambooth}, Custom Diffusion~\cite{multiconcept}, NeTI~\cite{neti}, and Textual Inversion~\cite{textual_inversion}. Note that the prompts used to generate these samples are not used in the RL fine-tuning step of InstructBooth (i.e., \textit{unseen} prompt).}
    \label{fig:unseen_examples_sup_2}
\end{figure*}

We provide additional generated examples comparing other existing approaches:  DreamBooth~\cite{dreambooth}, Custom Diffusion~\cite{multiconcept}, NeTI~\cite{neti}, and Textual Inversion~\cite{textual_inversion}. Figure~\ref{fig:seen_examples_sup} shows the generated samples using \textit{seen} prompt during RL fine-tuning. We further provide additional examples with \textit{unseen} prompts in Figure~\ref{fig:unseen_examples_sup_1} and~\ref{fig:unseen_examples_sup_2}.

\newpage
\section{Human Evaluation Details}~\label{sup:human_eval}
\begin{figure*}[b]
    \includegraphics[width=1.0\linewidth]{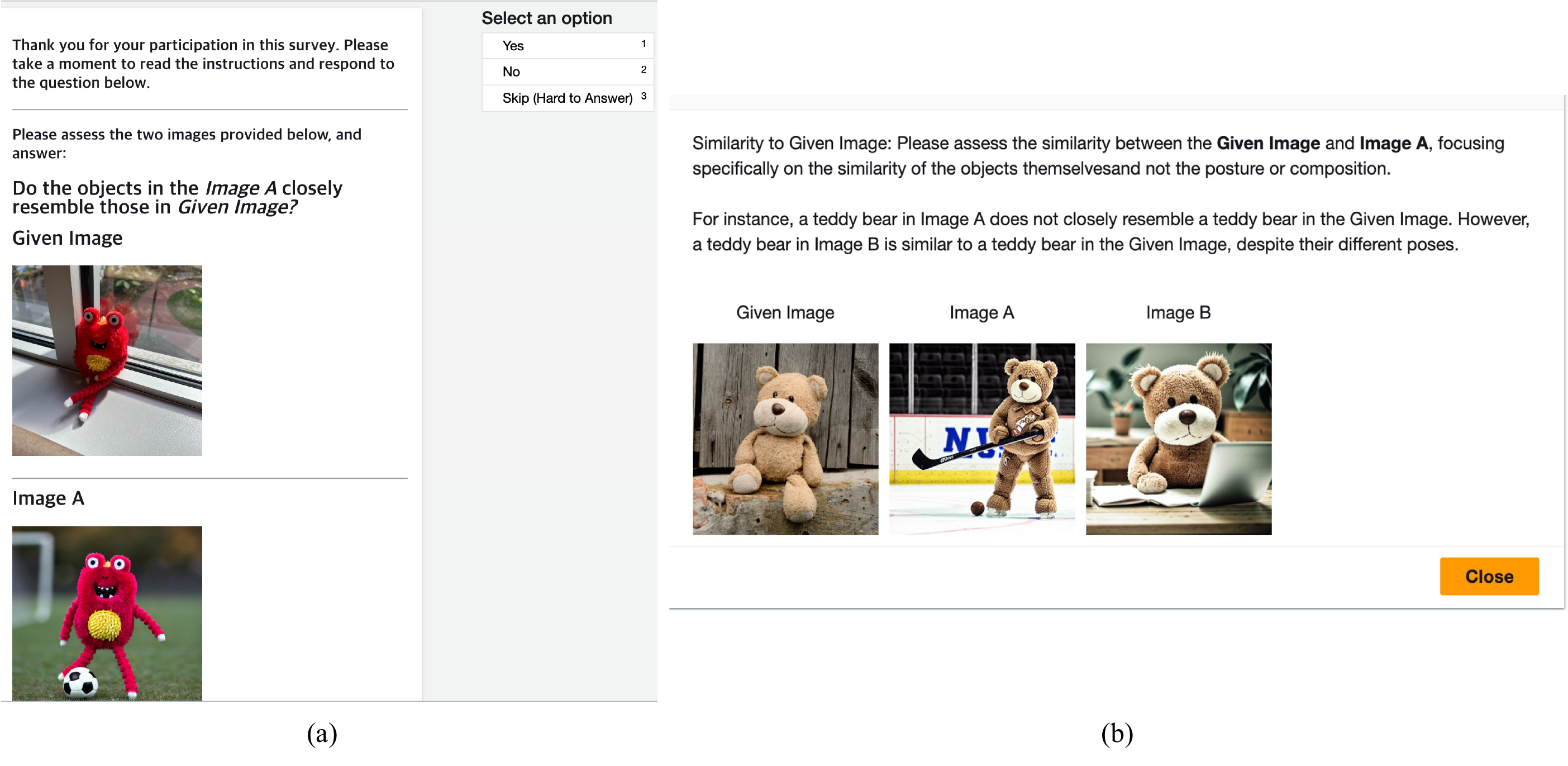}
    \vspace{-2em}
    \caption{(a) A screenshot of our human evaluation questionnaires to evaluate subject fidelity with (b) an instruction.}
    \label{fig:survey_sim}
\end{figure*}

Our human evaluation consists of three different parts: (i) subject fidelity evaluation, (ii) text fidelity evaluation, (iii) and overall quality evaluation. In (i), as shown in Figure~\ref{fig:survey_sim}, human raters are asked to answer yes or no to the following question: ``Do the objects in the Image A closely resemble those in Given Image?'' Importantly, human raters are asked to focus on perceptual resemblances without consideration of subjects' poses. We provide an example question on the instruction page to guide human raters to focus on perceptual resemblances.

\begin{figure*}[h]
    \includegraphics[width=1.0\linewidth]{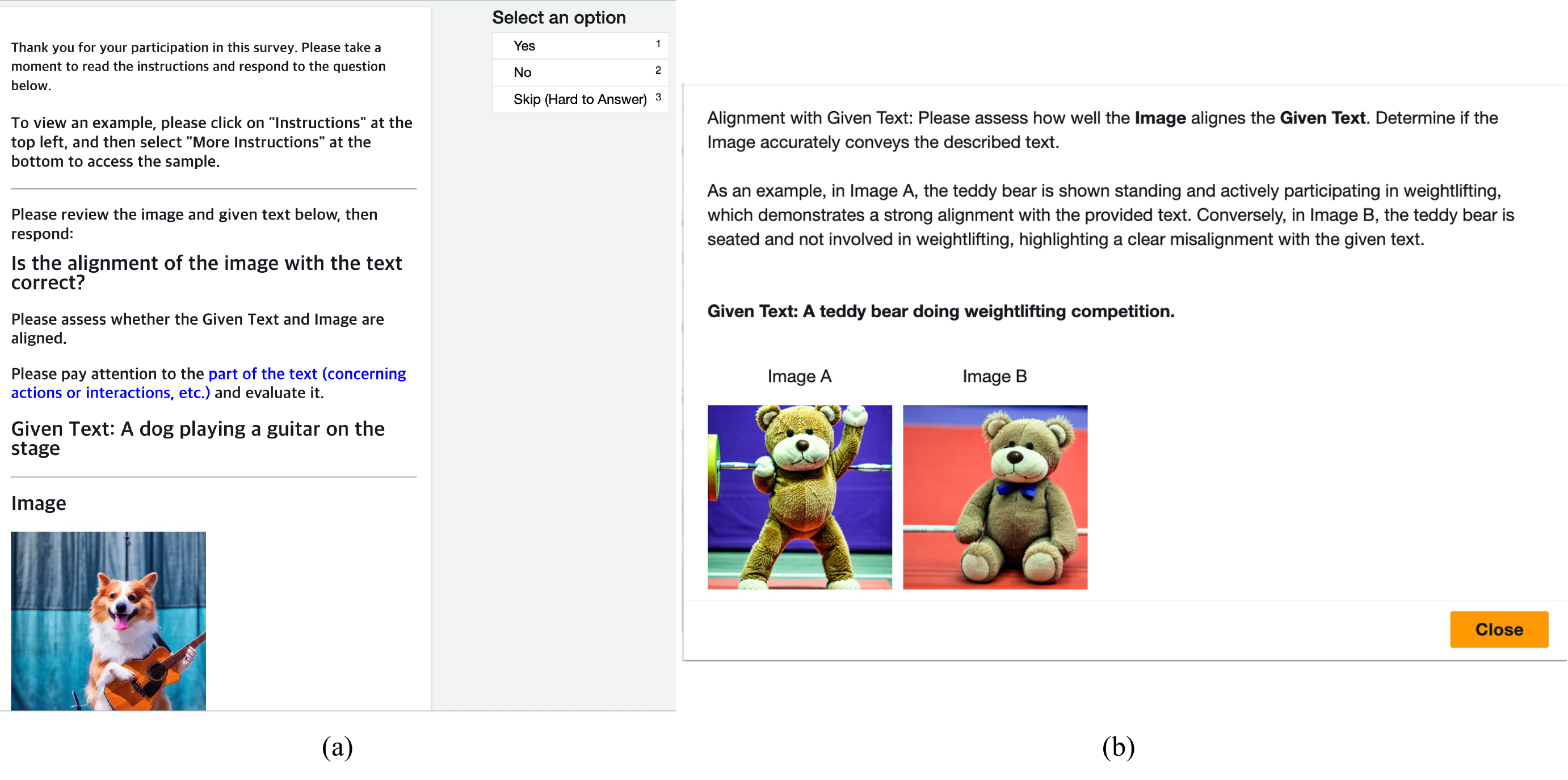}
    \vspace{-2em}
    \caption{(a) A screenshot of our human evaluation questionnaires to evaluate text fidelity with (b) an instruction.}
    \label{fig:survey_txt}
\end{figure*}

\begin{figure*}
    \centering
    \includegraphics[width=1.0\linewidth]{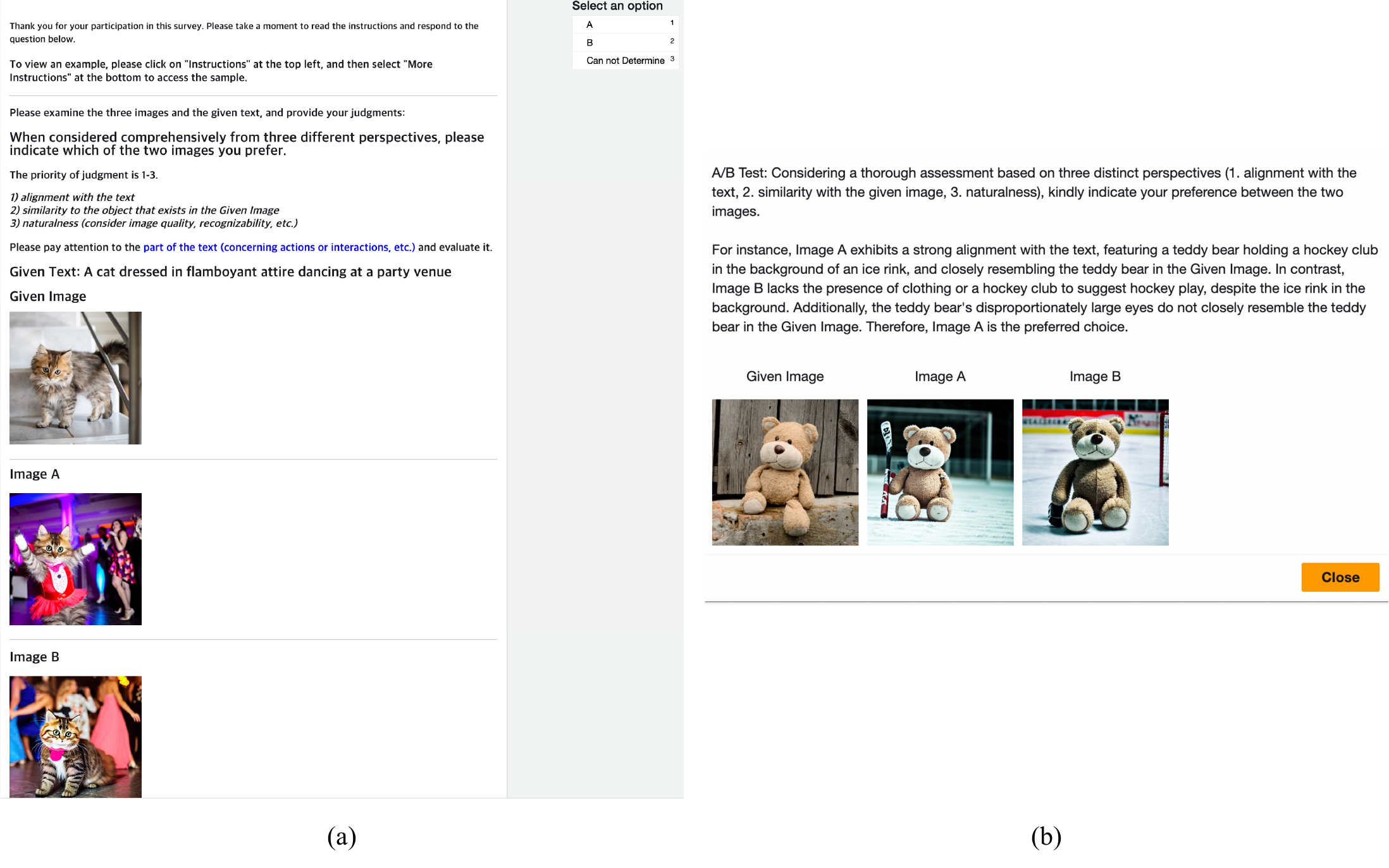}
    \vspace{-2em}
    \caption{(a) A screenshot of our human evaluation questionnaires to evaluate overall quality  with (b) an instruction.}
    \label{fig:survey_qual}
\end{figure*}

Further, we evaluate text fidelity, as shown in Figure~\ref{fig:survey_txt}, by asking human raters to answer the following question: ``Is the alignment of the image with the text correct?'' Like previous subject fidelity evaluations, we provide instructions to guide human raters to focus on image-text alignments. Lastly, we also conduct overall quality evaluation by asking human raters to answer the following question: ``When considered comprehensively from three different perspectives, please indicate which of the two images you prefer.'' In this question, we guide human raters to consider the following three perspectives with priority in order: text fidelity, subject fidelity, and naturalness (see Figure~\ref{fig:survey_qual}).

\section{Ablation study on detailed description in personalization}~\label{sup:ablation}
We propose a simple personalization technique that leverages detailed description during personalization. To verify its effectiveness, we conduct an ablation study on it. As mentioned in the main paper, we utilize Phryge (the Paris 2024 Olympic mascot) as our target subject and compare the model trained with a detailed prompt “a [*] triangular plushie” and that with a standard simple prompt “a [*] plushie’. For evaluation, we use the prompts of teddy bear in our dataset and generate 10 images from each prompt. We use DINO score~\cite{dino} to confirm that this technique helps the text-to-image model capture the features of the rare subject concretely. Table~\ref{tab:ablation_stdudy} shows that this simple technique improves the personalization ability.

\vspace{-1em}
\setlength{\tabcolsep}{5pt}
\begin{table}[h]
\caption{Comparison between the models trained with detailed prompt and with a standard simple prompt.}
\vskip 0.15in
\begin{center}
\label{tab:ablation_stdudy}
\begin{tabular}{lcccr}
\toprule
Method & DINO \\
\midrule
With detailed description & \textbf{0.524} \\
Without detailed description & 0.489\\
\bottomrule
\end{tabular}
\end{center}
\vskip -0.1in
\end{table}

\section{DreamBench Evaluation Details}~\label{sup:dreambench}
DreamBench~\cite{dreambooth} comprises $30$ subject images with $21$ inanimate subjects (e.g., bear plushie, can, clock, monster toy) and $9$ live subjects (e.g., dogs and cats). The evaluation prompts for the inanimate subject category consist of 20 recontextualization prompts (e.g., `` [*] [class noun] in the jungle'') and $5$ property modification prompts (e.g., `` a shiny [*] [class noun]''). Note that [*] denotes a unique identifier. For the live subjects, there are $10$ recontextualization prompts, $10$ accessorization prompts (e.g., `` [*] [class noun] wearing a red hat''), and $5$ property modification prompts. For more details, see the following link\footnote{\href{https://github.com/google/dreambooth/blob/main/dataset}{https://github.com/google/dreambooth/blob/main/dataset}}.

Importantly, we do not use DreamBench prompts directly during RL fine-tuning for a fair comparison. Instead, we use the following two prompts for fine-tuning with inanimate subjects:
\begin{itemize}
    \item ``A [*] [class noun] with a church in the background''
    \item ``A [*] [class noun] on top of blue sofa'',
\end{itemize}

\begin{figure*}
    \centering
    \includegraphics[width=1.0\linewidth]{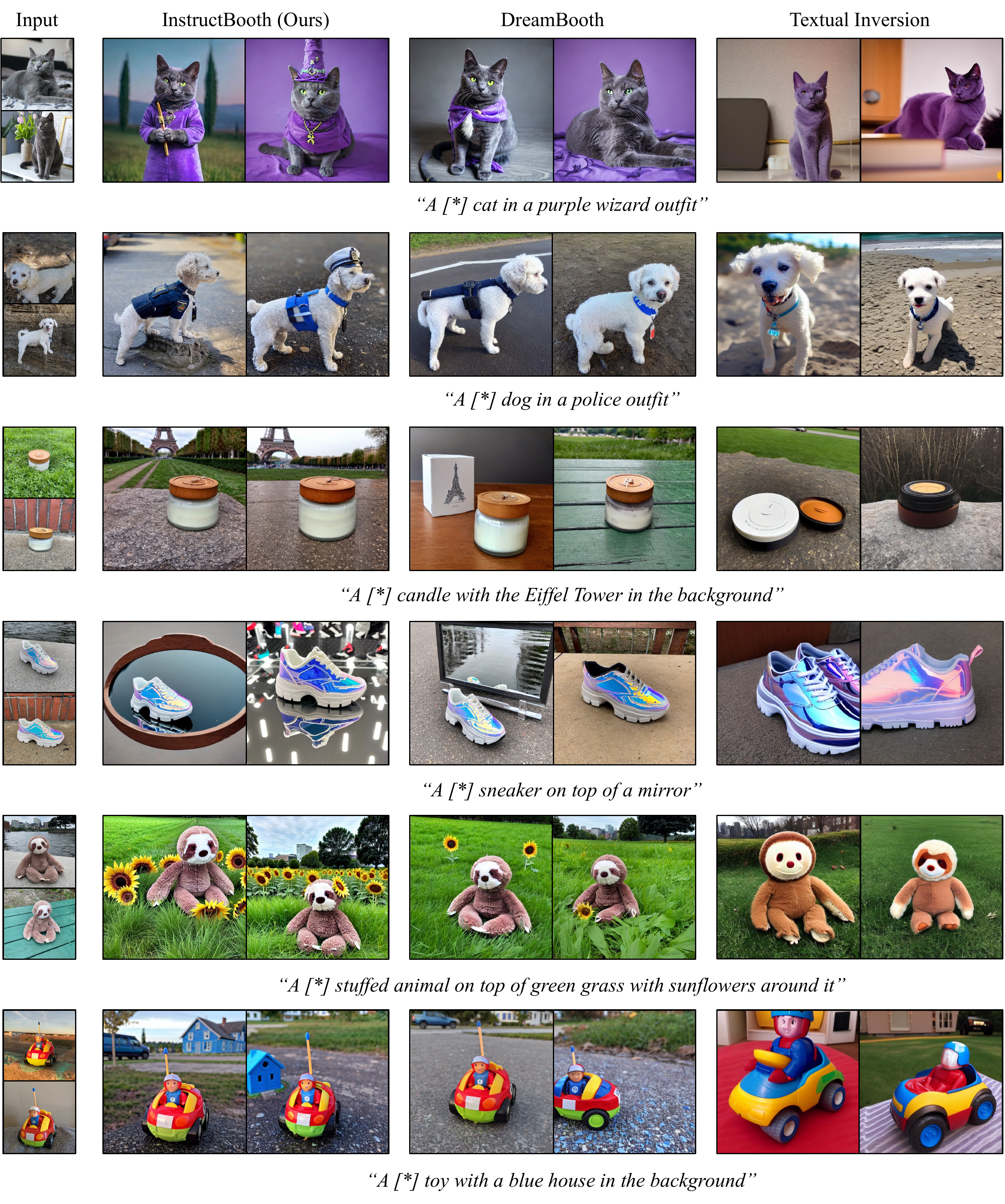}
    \vspace{-2em}
    \caption{Samples generated by InstructBooth, DreamBooth~\cite{dreambooth} and Textual Inversion~\cite{textual_inversion} using DreamBench's prompts.}
    \label{fig:dreambench}
\end{figure*}

For live subjects, we use two following text prompts for RL fine-tuning:
\begin{itemize}
    \item ``A [*] [class noun] in a nurse outfit''
    \item ``A [*] [class noun] on top of blue sofa''
\end{itemize}
In addition to the evaluation scores reported in the main paper, we provide samples generated by InstructBooth, DreamBooth and Textual Inversion~\cite{textual_inversion} in Figure~\ref{fig:dreambench}.

\end{document}